\documentclass[10pt,twocolumn,letterpaper]{article}

\usepackage[pagenumbers]{cvpr} % To force page numbers, e.g. for an arXiv version

\usepackage{lipsum}  
\usepackage{pifont}% http://ctan.org/pkg/pifont
\newcommand{\cmark}{\ding{51}}%
\newcommand{\xmark}{\ding{55}}%
\usepackage{multirow}
\usepackage{makecell}
\usepackage{algorithm}
\usepackage{algpseudocode}

\usepackage{diagbox}  % For diagonal line in the cell

\usepackage{mwe}

\usepackage{tikz}

\usepackage{setspace}
\renewcommand{\arraystretch}{1} % Adjust the number to increase space (1.5 is 50% more space)

\usepackage{hhline}
\usepackage{siunitx,booktabs}

\usepackage{amsmath}
\DeclareMathOperator*{\argmax}{arg\,max}
	
\usepackage{color, colortbl}
\definecolor{LightGray}{gray}{0.9}
\definecolor{darkerGreen}{rgb}{0,0.5,0}
\usepackage{graphicx}
\usepackage{wrapfig}
\usepackage{amsmath} 
\usepackage{tabularx}
\usepackage{arydshln}
\usepackage{float}
\usepackage[export]{adjustbox}
\usepackage{multirow}
\usepackage{dsfont}
\usepackage{amssymb}% http://ctan.org/pkg/amssymb
\usepackage{pifont}% http://ctan.org/pkg/pifont
\renewcommand{\xmark}{\ding{55}}%

\usepackage{mathtools}
\newcommand{\cleq}{\stackrel{\mathclap{\normalfont\mbox{c}}}{\leq}}

\definecolor{cvprblue}{rgb}{0.21,0.49,0.74}
\usepackage[pagebackref,breaklinks,colorlinks,citecolor=cvprblue]{hyperref}

\def\paperID{12991} % *** Enter the Paper ID here
\def\confName{CVPR}
\def\confYear{2024}

\title{On the test-time zero-shot generalization of vision-language models: \\ Do we really need prompt learning?}

\author{Maxime Zanella\thanks{Corresponding author: maxime.zanella@uclouvain.be\\This work was partly supported by the Walloon Region
(Service Public de Wallonie Recherche, Belgium) under grant
n°2010235 (ARIAC by DigitalWallonia4.ai).}\\
UCLouvain \quad UMons 
\and Ismail Ben Ayed
\\ \'ETS Montr\'eal \and 
{\tt code: \url{https://github.com/MaxZanella/MTA}}
}

\begin{document}
\maketitle
\begin{abstract}
The development of large vision-language models, notably CLIP, has catalyzed research into effective adaptation techniques, with a particular focus on soft prompt tuning. Conjointly, test-time augmentation, which utilizes multiple augmented views of a single image to enhance zero-shot generalization, is emerging as a significant area of interest. This has predominantly directed research efforts toward test-time prompt tuning. In contrast, we introduce a robust \textbf{M}eanShift for \textbf{T}est-time \textbf{A}ugmentation (MTA), which surpasses prompt-based methods without requiring this intensive training procedure. This positions MTA as an ideal solution for both standalone and API-based applications. Additionally, our method does not rely on ad hoc rules (e.g., confidence threshold) used in some previous test-time augmentation techniques to filter the augmented views. Instead, MTA incorporates a quality assessment variable for each view directly into its optimization process, termed as the inlierness score. This score is jointly optimized with a density mode seeking process, leading to an efficient training- and hyperparameter-free approach. We extensively benchmark our method on 15 datasets and demonstrate MTA's superiority and computational efficiency. Deployed easily as plug-and-play module on top of zero-shot models and state-of-the-art few-shot methods, MTA shows systematic and consistent improvements. %Code is available at \url{https://github.com/MaxZanella/MTA}. 
\end{abstract}
\section{Introduction}
Vision-language models, pretrained on vast sets of image-text pairs, have emerged as powerful tools for learning cross-modal representations~\cite{clip, align, filip, lit, li2022grounded, vlm_review}. The joint feature space of visual and textual features enables zero-shot recognition, without any task-specific data. For instance, given a set of candidate classes, one can create textual descriptions, with the so-called prompt~\cite{liu2023pre}, ${\mathbf p}_k$~=~`'\texttt{a photo of a} [$\texttt{class}_k$]", and get its corresponding embedding representation ${\mathbf t}_k = \theta_t ({\mathbf p}_k)$ with the language encoder. Similarly, an image ${\mathbf x}$ is projected in the same embedding space ${\mathbf f} = \theta_v ({\mathbf x})$ using the visual encoder. Then, one can classify this image by measuring the similarity between these two encoded modalities and predicting the class corresponding to the most similar embedding, $\hat{k} = \argmax_{k} {\mathbf f}^t {\mathbf t}_k$.

Despite their impressive capabilities, these models still encounter substantial challenges and may yield unsatisfactory responses in complex situations~\cite{goh2021multimodal, clip}. These issues are particularly pronounced when confronted with the pragmatic constraints of real-world scenarios, where labeled data can be scarce (i.e., few-shot scenarios~\cite{song2023comprehensive}) or completely absent (i.e., zero-shot scenarios~\cite{lampert2009learning}), thus limiting their broader usage. Consequently, there has been a growing interest in enhancing the test-time generalization faculties of these vision-language models~\cite{tpt, difftpt, ma2023swapprompt, CupL, chils, hierarchy}.\\  Empirical findings indicating that improved textual descriptions can positively impact zero-shot predictions~\cite{clip} have sparked interest in refining prompt quality for downstream tasks. Originating in the NLP community~\cite{ShinEMNLP2020,JiangACL2020,Hong2021NAACL}, soft prompt learning, which utilizes learnable continuous tokens as input~\cite{prompt_tuning}, has rapidly gained popularity. Building on this momentum, CoOp~\cite{coop} stands out as the seminal work for prompt tuning in vision-language models. Since then, prompt tuning has appeared as the prominent approach for adapting vision-language models~\cite{vlm_review} across both unsupervised~\cite{upl, tpt, difftpt} and few-shot scenarios~\cite{coop, cocoop, proda, variational, kgcoop, lasp, prograd, plot}.

In parallel, test-time augmentation, which has been extensively used in the computer vision community~\cite{memo, survey_tta, liang2023comprehensive}, is now emerging in the vision-language field, with a focus on prompt tuning~\cite{tpt, difftpt, ma2023swapprompt}. Instead of exploiting a single image ${\mathbf x}$, {\em test-time prompt tuning} techniques leverage multiple embeddings $({\mathbf f}_p)_{1 \leq p \leq N}$, each derived from a different augmented view $({\mathbf x}_p)_{1 \leq p \leq N}$ of the same original image ${\mathbf x}$. Afterwards, the prompt is optimized by forcing consistency of the predictions among these different views~\cite{tpt}. The final classification step is then performed by computing the similarity between the original image encoding and the {\em optimized} textual embedding ${\mathbf t}_k^*$, $\hat{k} = \argmax_{k} {\mathbf f}^t {\mathbf t}_k^*$. These novel research directions underscore the increasing attention in enhancing these models' robustness, especially in zero-shot scenarios, through data augmentation at test-time. Alongside this expanding literature, we ask the following question: {\em Can we improve the image representation ${\mathbf f}$ directly in the embedding space, achieving superior results in a way that is more efficient than prompt tuning?}

Concurrently, there has been a surge in the use of proprietary and closed APIs that encapsulate advanced machine learning functionalities, often termed {\em black boxes} due to their limited transparency, offering little insight into their internal mechanisms or architectures. Yet, they are crucial in executing a wide spectrum of tasks in vision and NLP, introducing new challenges in model adaptation~\cite{solaiman2023gradient}. The field of NLP, in particular, has seen an emerging literature on few-shot adaptation of black-box models~\cite{colombo2023transductive}, driven by the reality that large-scale models (e.g., GPT family~\cite{brown2020language, openai2023gpt4v}, Palm~\cite{chowdhery2022palm}) are only accessible via APIs and their pretrained weights are not publicly available. Optimizing prompts, which necessitates gradient computation from output back to input, a memory-intensive and time-consuming process, is impractical in the context of API-reliant applications. In contrast, our approach does not require extra assumptions about the model's internal states or architecture, making it suitable for black-box applications.

\paragraph{Contributions.} In this work, we introduce a robust multi-modal \textbf{M}eanShift \textbf{T}est-time \textbf{A}ugmentation (MTA), which enhances the zero-shot generalization of CLIP models, leveraging different augmented views of a given image. Unlike current prompt tuning solutions, which rely on heavy training procedures and {\em ad hoc} thresholds to discard degenerated views, MTA uses only the final embedding state and directly integrates an {\em inlierness} assessment of the augmented views into its optimization process.
Our objective function is efficiently solvable using iterative block coordinate descent updates, and relaxes the need for training the model's parameters or prompts.
Empirically, we demonstrate that MTA surpasses state-of-the-art prompt-tuning alternatives, while being time and memory efficient. Our key contributions are as follows:
\begin{enumerate}
\item We propose a robust MeanShift formulation, which automatically manages augmented views in test-time augmentation scenarios by optimizing {\em inlierness} variables. 
Used as a versatile {\em plug-and-play} tool, MTA improves the zero-shot performances of various models on a large variety of classification tasks, without any hyperparameter tuning.
\item We report comprehensive evaluations and comparisons to the existing test-time prompt tuning techniques on 15 datasets, showing MTA's highly competitive performances, although it operates in limited access (i.e., final embedding) and training-free mode. This makes MTA suitable for both standalone and API-based applications.
\item Deployed easily atop current state-of-the-art few-shot learning methods, MTA brings consistent improvements, a benefit not observed with test-time prompt tuning. 
\end{enumerate}

\section{Related works}
\paragraph{Vision-language models adaptation.} Large scale vision-language models have shown excellent results in several vision tasks~\cite{vlm_review}. This success has created interest in developing adaptation techniques that capitalize their general knowledge~\cite{fine_tuning_clip}. Among these, prompt tuning~\cite{prompt_tuning} has emerged as the primary method for adapting CLIP-like models, at test-time based on data augmentations~\cite{tpt, difftpt, ma2023swapprompt} or with few labeled samples~\cite{coop, cocoop, proda, variational, kgcoop, lasp, prograd, plot}. CoOp~\cite{coop} optimizes learnable continuous tokens attached to the class name, while CoCoOp~\cite{cocoop} trains a neural network to generate instance-conditioned tokens based on the image. Further efforts include ProGrad~\cite{prograd}, which guides prompts toward predefined handcrafted ones based on gradients, whereas PLOT~\cite{plot} aligns learned prompts with finer-grained visual features via an optimal transport formulation.
Beyond soft prompt tuning, other strategies involve using hierarchical word structures to create more semantically refined class descriptions~\cite{hierarchy, chils}, or exploiting other large scale models to generate more detailed prompts~\cite{cafo, sus, CupL} or new images by diffusion mechanisms~\cite{cafo, difftpt}.\\
Contrastingly, methods such as CLIP-Adapter~\cite{clip-adapter} offer an alternative strategy by learning feature adapters. However, there has been limited effort in developing black-box methods~\cite{blackbox}, which can effectively capitalize the knowledge of these models while only accessing their final embedding state. Examples include zero-shot prediction with parameter-free plug-in attention~\cite{calip}, or few-shot settings with Tip-Adapter~\cite{tip-adapter} using a cache model.\\ 
Our experiments demonstrate that our robust MeanShift algorithm significantly enhances the performances in zero-shot scenarios, without relying on soft prompt tuning, while respecting the black-box constraints. Additionally, we report increased performances when applied atop of various aforementioned few-shot methods, without requiring further training or hyperparameter tuning.
\paragraph{Test-time augmentation.} Data augmentation during training is widely recognized for its capacity to enhance model robustness~\cite{augmix, hendrycks2021many}. Also, its utility extends to test-time applications~\cite{memo, survey_tta, liang2023comprehensive}. In particular, test-time augmentation can be used on a single image~\cite{memo} to adapt models with an entropy minimization term. The latter is often used in the context of unsupervised adaptation~\cite{shot, tent}, but is deployed differently in this augmentation setting, enforcing consistent predictions across the various augmented views. This idea is further developed for vision-language models with test-time prompt tuning (TPT)~\cite{tpt}, where a prompt is optimized to make consistent predictions among light augmentations inspired by Augmix~\cite{augmix}. DiffTPT~\cite{difftpt} builds up on this work by adding generated images from Stable Diffusion~\cite{stable_diffusion} to acquire more diverse views. Both works show improvements when selecting a subset of the augmented views. Specifically, TPT utilizes only the 10\% most confident views, and DiffTPT measures the similarity with the original image, keeping the unconfident but correctly classified augmentations. We also demonstrate that filtering the augmented views can substantially improve test-time augmentation techniques. Additionally, our method does not rely on arbitrary hard thresholds or rules as in TPT and DiffTPT; instead, we directly integrate the weighting of the augmented views in our optimization procedure thanks to {\em inlierness} variables.

% Independently of works on finding [ref] or learning ~\cite{kim2020learning} the best data augmentation process, our work centers on how to aggregate and treat these augmentations, which has already been explored in the past, for example, by averaging the resulting different predictions ~\cite{survey_tta} or learning it ~\cite{aggregate_tta}. On the other hand, our robust MeanShift seeks for a mode in a dense area, leveraging consistent features across the several augmentations while naturally avoiding inconsistent ones.

\section{Robust multi-modal MeanShift}
\label{sec:method}

Similarly to the test-time generalization setting recently introduced in TPT~\cite{tpt}, let us assume that we are given a set of image samples $({\mathbf x}_p)_{1 \leq p \leq N}$, which correspond to $N$ distinct augmented views of a given test sample ${\mathbf x}$. It is important to note that our method is applicable on top of any type of 
augmentations. Let ${\mathbf f}_p = \theta_v ({\mathbf x}_p)$ denote the vision-encoded feature embedding corresponding to augmented sample ${\mathbf x}_p$, $\theta_v$ being 
the vision encoder of the CLIP pre-trained model. A straightforward way to use the ensemble of augmentations is to perform the zero-shot prediction based on their mean embedding, 
thereby giving exactly the same importance to all augmented samples, independently of the structure of the data. However, the augmentations may include degenerated views, which correspond to {\em outliers}, e.g., in the form of isolated data points or small regions with little structure within the feature space. Such outliers may bias global statistics like the mean. 

\subsection{Formulation}
We hypothesize that {\em robust statistics} like the modes of the density of the set of augmented views could provide better representations. 
Augmented views presenting major characteristics of the concept to be recognized are likely to be projected close to the original image and close to each 
others. This motivates our formulation, which could be viewed as a novel robust and multi-modal extension of the popular MeanShift algorithm \cite{comaniciu1999mean}, an 
unsupervised procedure for finding the modes of the distribution of a given set of samples. In our case, we explicitly model and handle the potential presence of outliers in the estimation of the kernel density of the set of feature vectors $({\mathbf f}_p)_{1 \leq p \leq N}$. To do so, we introduce a latent assignment vector $\mathbf{y} = (y_p)_{1 \leq p \leq N} \in \Delta^{N-1}$, with $\Delta^{N-1} = \{{\mathbf y} \in [0, 1]^N \; | \; {\mathbf 1}^t {\mathbf y}   = 1 \}$ the probability simplex, and propose to minimize the 
following objective function:
\begin{align}
\label{eq:robust_mean_shift}
    & \quad \quad  \min_{{\mathbf m}, {\mathbf y}} {\cal L} ({\mathbf m}, {\mathbf y}) \quad \mbox{s.t.} \quad {\mathbf y} \in \Delta^{N-1}  \quad \mbox{with} \nonumber \\  
     & {\cal L} ({\mathbf m}, {\mathbf y}) = - \sum_{p=1}^N y_p K({\mathbf f}_p - {\mathbf m}) - \frac{\lambda}{2} \sum_{p, q} w_{p,q} y_p y_q   \nonumber \\
     &- \lambda_{\mathbf{y}} H({\mathbf y})
\end{align}
In the following, we describe the notations occurring in our model in Eq. (\ref{eq:robust_mean_shift}), as well as the effect of each of its terms:

\paragraph{Robust KDE (first term)} $K$ is a kernel function measuring a robust affinity between ${\mathbf f}_p$ and ${\mathbf m}$, e.g., a Gaussian kernel \cite{CarreiraPerpin2007GaussianMI}: $K({\mathbf f}_p - {\mathbf m}) \propto \exp (-\|{\mathbf f}_p - {\mathbf m}\|^2/h^2)$, where $h$ is the kernel bandwidth.
When variables $y_p$ are fixed to $1$ $\forall p$, the first term reduces to the kernel density estimate (KDE) of the distribution of features at point $\mathbf{m}$.
Clearly, minimizing this term w.r.t $\mathbf{m}$ yields the standard MeanShift algorithm for finding the mode of the density (i.e., the point maximizing it).   
In our model, the additional latent variable $y_p$ evaluates the \textit{inlierness} of the $p^{\mbox{{\small th}}}$ augmented view, 
i.e., the model's belief in ${\mathbf f}_p$ being an inlier or not within the whole set of augmented-view embeddings $({\mathbf f}_p)_{1 \leq p \leq N}$. 
Score $y_p \in [0, 1]$ is high when the model considers the $p^{\mbox{{\small th}}}$ sample as an inlier, enabling it to contribute more in the KDE evaluation in (\ref{eq:robust_mean_shift}),  and small (closer to $0$) otherwise.

\paragraph{Text-knowledge guided quadratic term (second term)} 
This term encourages samples with nearby text-based zero-shot predictions
to have similar {\em inlierness} scores $y_p$. Specifically, we construct the pairwise affinities $w_{p,q}$ in the second term of (\ref{eq:robust_mean_shift}) from both the text and vision embeddings as follows. Let ${\mathbf s}_p \in \mathbb R^K$ denotes the text-driven softmax prediction based on the zero-shot text embedding for the $p^{\mbox{{\small th}}}$ sample, i.e., the $k^{\mbox{{\small th}}}$ component of ${\mathbf s}_p$ is given by: 
\begin{equation}
s_{p,k} = \frac{\exp l_{p,k}}{\sum_{j=1}^K \exp l_{p,j}}; \, l_{p,k} = \tau {\mathbf f}_p^t {\mathbf t}_k
\end{equation}
where $\tau$ is the temperature scaling parameter of the CLIP model. Affinities $w_{p,q}$ are given by:   
\begin{equation}
\label{eq:affinity_term}
w_{p,q} =  {\mathbf s}^t_p {\mathbf s}_q
\end{equation}

\paragraph{The Shannon entropy (third term)} $H({\mathbf y})$ is the Shannon entropy defined over simplex variables as follows: 
\begin{equation}
H({\mathbf y}) = - \sum_{p=1}^N y_p \ln y_p
\end{equation}
This term acts as a barrier function, forcing latent variable ${\mathbf y}$ to stay within the probability simplex. Furthermore, this 
entropic regularizer is necessary to avoid a trivial solution minimizing the quadratic term (i.e., $y_p = 1$ ~ $\exists p$.), as it pushes the solution toward the middle of the simplex (i.e., $y_p =1/N \,  ~ \forall p$). 

\subsection{Block coordinate descent optimization}
Our objective in \eqref{eq:robust_mean_shift} depends on two types of variables: ${\mathbf m}$ and ${\mathbf y}$. Therefore, we proceed with block-coordinate descent alternating two sub-steps: one optimizing \eqref{eq:robust_mean_shift} w.r.t $\mathbf{y}$ and keeping density modes $\mathbf{m}$ fixed, while the other minimizes \eqref{eq:robust_mean_shift} w.r.t $\mathbf{m}$ with the {\em inlierness} variables fixed. 

\paragraph{Optimization w.r.t $\mathbf y$ via the concave-convex procedure}
 When ${\mathbf m}$ is fixed, our objective ${\cal L} ({\mathbf y}, {\mathbf m})$ could be minimized efficiently w.r.t ${\mathbf y}$ using the Concave-Convex Procedure (CCCP) \cite{cccp}, with convergence guarantee. 
At each iteration, we update the current solution ${\mathbf y}^{(n)}$ as the minimum of a tight upper bound on ${\cal L}$, which ensures the objective does not increase.
For the sum of concave and convex functions, as for our sub-problem, the CCCP replaces the concave part by its linear first-order approximation at the current solution, which is a tight upper bound, while keeping the convex part.
For \eqref{eq:robust_mean_shift}, the quadratic term could be written as
$\mathbf{y}^t W \mathbf{y}$, with $W = [w_{i,j}]$. It is easy to see that this term is concave as affinity matrix $W$ is positive semi-definite, whereas the remaining part of $\mathcal{L}$ is convex. Therefore, we replace this quadratic term by $\mathbf{y}^t W^t \mathbf{y}^{(n)}$, obtaining, up to an additive constant, the following tight bound:
\begin{equation}
\label{L-bound}
{\cal L} ({\mathbf y}, {\mathbf m}) \cleq - \sum_{p=1}^N y_p K({\mathbf f}_p - {\mathbf m}) - \lambda \mathbf{y}^t W^t \mathbf{y}^{(n)} - \lambda_{\mathbf{y}} H({\mathbf y})
\end{equation}
Solving the Karush-Kuhn-Tucker (KKT) conditions for minimizing bound
\eqref{L-bound}, s.t. simplex constraint ${\mathbf y} \in \Delta^{N-1}$, gives the following updates for $\mathbf{y}$:
\begin{align}
	\label{final-updates-y}
  y_p^{(n+1)} = \frac{\exp \left( (K({\mathbf f}_p - {\mathbf m}) + \lambda \sum_{q = 1}^N w_{p,q} y_q^{(n)} )/ \lambda_{\mathbf{y}} \right)}{\sum_{j=1}^N \exp \left ((K({\mathbf f}_j - {\mathbf m}) + \lambda \sum_{q = 1}^N w_{j,q} y_q^{(n)} )/ \lambda_{\mathbf{y}}\right)}
\end{align}
which have to be iterated until convergence. The complete derivation of Eq. \eqref{final-updates-y} is provided in Appendix \ref{appendix:proof_update_y}.

\paragraph{Optimization w.r.t $\mathbf m$ via fixed-point iterations}
This sub-step fixes ${\mathbf y}$, and minimizes the objective in (\ref{eq:robust_mean_shift}) w.r.t the density modes $\mathbf{m}$. Setting the gradient of ${\cal L}$ w.r.t 
$\mathbf{m}$ to 0 yields the following necessary condition for a minimum, which takes the form of a fixed-point equation:
\begin{equation}
\label{eq:condition}
    \mathbf{m} - g(\mathbf{m}) = 0; \quad g(\mathbf{m}) = \frac{\sum_{p = 1}^{N}  y_p K(\mathbf{f}_p - \mathbf{m}) \mathbf{f}_p}{\sum_{p = 1}^N y_p  K(\mathbf{f}_p - \mathbf{m})} 
\end{equation}
The solution to (\ref{eq:condition}) could be obtained by the following fixed-point iterations: 
\begin{equation}
\label{fixed-point-iteration-updates}
\mathbf{m}^{l+1} = g(\mathbf{m}^{l}) = \frac{\sum_{p = 1}^{N}  y_p K(\mathbf{f}_p - \mathbf{m}^l) \mathbf{f}_p}{\sum_{p = 1}^N y_p  K(\mathbf{f}_p - \mathbf{m}^l)}  
\end{equation}
This yields a Cauchy sequence $\{{\mathbf m}^{l}\}_{l \in \mathbb{N}}$, which converges to a unique value:
$\mathbf{m}^{*} = \lim_{l \xrightarrow{} \infty} {\mathbf m}^{l+1} = \lim_{l \xrightarrow{} \infty} g(\mathbf{m}^{l}) = g (\lim_{l \xrightarrow{} \infty} \mathbf{m}^{l}) = g(\mathbf{m}^{*})$, and $\mathbf{m}^{*}$ is the unique solution of the fixed-point in \eqref{eq:condition} (Appendix \ref{appendix:cauchy_sequence}).

\paragraph{Final prediction} The class prediction is computed by the cosine similarity between the mode minimizing our objective 
in Eq. \eqref{eq:robust_mean_shift}, i.e., $\mathbf{m}^{*}$ obtained at convergence, and the encoded prompts of each class k, i.e., $\mathbf{t}_k$:
\[\hat{k} = \argmax_{k} ~({\mathbf m}^{*})^t {\mathbf t}_k \]

\begin{figure}[t!]
    \centering
    \begin{subfigure}{\columnwidth}
        \centering
        \includegraphics[width=0.88\linewidth]{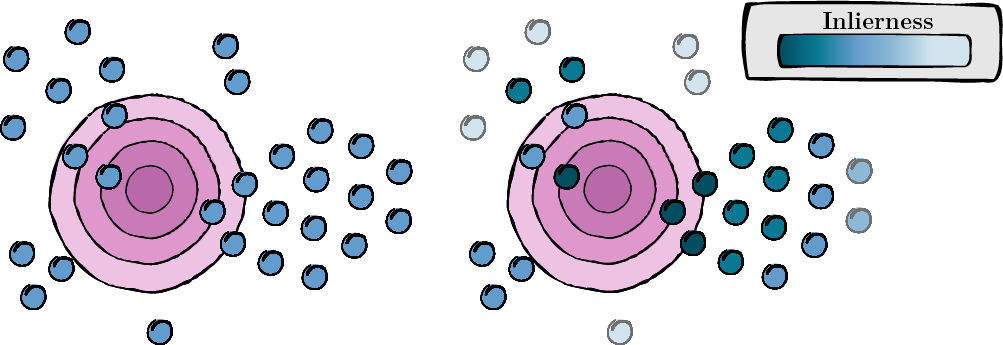} % Adjust width as needed
        \caption{\textbf{\textit{Inlierness scores distribution.}} Blue gradation represents the {\em inlierness} score: the darker, the higher.}
        \label{fig:interpretation_inlier}
    \end{subfigure}
    
    \begin{subfigure}{\columnwidth}
        \centering
        \includegraphics[width=0.88\linewidth]{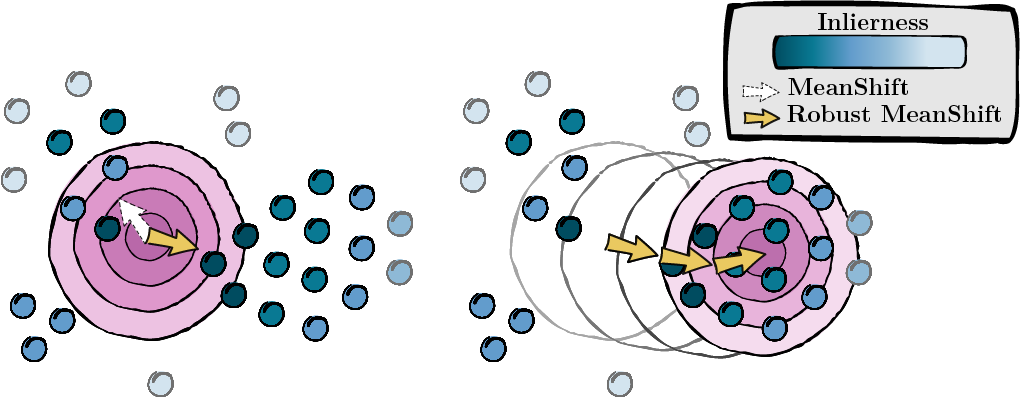} % Adjust width as needed
        \caption{\textbf{\textit{Inlierness-density mode seeking.}} The white arrow represents the update direction of the traditional MeanShift; instead our robust MeanShift follows the orange ones.}
        \label{fig:interpretation_mode}
    \end{subfigure}

    \caption{Interpretation of Eqs. \eqref{final-updates-y} and \eqref{fixed-point-iteration-updates} in (a) and (b) respectively. Our robust MeanShift alternatively solves these 2 equations until convergence. A pseudocode is available in Appendix \ref{appendix:further_details}.}
    \label{fig:interpretation}
\end{figure}

\begin{table*}[t!]
\caption{Zero-shot methods on ImageNet datasets. We use "\texttt{a photo of a} [$\texttt{class}_k$]" as prompt. Majority vote of the 80 handcrafted prompts of \cite{clip} is used for final prediction when Ensemble is specified. CoOp is a pretrained prompt on 16-shots ImageNet. We highlight the best and second best results by {\bf bolding} and \underline{underlining} them, respectively. \cmark \hspace{0.05cm} for training-free methods at test-time, \xmark \hspace{0.05cm} otherwise.} \label{tab:zero_shot_imagenet}
\sisetup{table-format=-1.2}   % 2 decimals, leave 
\centering
\resizebox{0.75\linewidth}{!}{
   \begin{tabular}{llcccccc}
\toprule 
\multicolumn{1}{l}{\bf Method}  
&\multicolumn{1}{c}{}
&\multicolumn{1}{c}{\makecell{\bf ImageNet}}
&\multicolumn{1}{c}{\makecell{\bf -A}} 
&\multicolumn{1}{c}{\makecell{\bf -V2}} 
&\multicolumn{1}{c}{\makecell{\bf -R}}
&\multicolumn{1}{c}{\makecell{\bf -Sketch}}
&\multicolumn{1}{c}{\bf Average} 
\\ \midrule
CLIP~\cite{clip}   
& \cmark       & 66.73  & 47.87 & 60.86 & 73.98 & 46.06 &  59.11   \\
CLIP + Ensemble~\cite{clip}   
& \cmark          & 68.38 & 49.95 & 62.1 & \underline{77.38} & 47.96 & 61.15  \\
\midrule
TPT~\cite{tpt} & \xmark          & 68.94 & 54.63 & 63.41 & 77.04 & 47.97  & 62.40  \\
MTA (Ours)    &    \cmark    & \underline{69.29} & \underline{57.41} & \underline{63.61} & 76.92 & \underline{48.58} & \underline{63.16} \\
\rowcolor{LightGray} MTA + Ensemble (Ours)   &    \cmark    & \textbf{70.08} & \textbf{58.06} & \textbf{64.24} & \textbf{78.33} & \textbf{49.61} & \textbf{64.06} \\
\midrule 
\midrule 
CoOp~\cite{coop} %from TPT paper
     & \cmark & 71.51 & 49.71 & 64.20 & 75.21 & 47.99    & 61.72 \\  
\midrule
TPT + CoOp~\cite{tpt}    & \xmark  & \underline{73.61} & \underline{57.85} & \underline{66.69} & \underline{77.99} & \underline{49.59}     & \underline{65.14} \\
\rowcolor{LightGray} MTA + CoOp (Ours) &  \cmark  & \textbf{73.99} & \textbf{59.29} & \textbf{66.97} & \textbf{78.2} & \textbf{49.96} & \textbf{65.68} \\

      \bottomrule
   \end{tabular}}
\end{table*}

\begin{table*}[t!]
\caption{Zero-shot methods on 10 fine-grained classification datasets. We highlight the best result by {\bf bolding}. +E. means that majority vote with the 80 handcrafted prompts of \cite{clip} is used for final prediction.}
\label{tab:zero_shot_datasets}
\centering
\resizebox{\textwidth}{!}{
\begin{tabular}{lccccccccccc}
\toprule
Method & SUN397 & Aircraft & EuroSAT & Cars & Food101 & Pets &  Flower102 & Caltech101 & DTD & UCF101 & Average
\\ \midrule 
CLIP~\cite{clip}

& 62.59 & 23.67 & 42.01 & 65.48 & 83.65 & 88.25 & 67.44 & 93.35 & 44.27 & 65.13 & 63.58 \\
CLIP + E.~\cite{clip} & 66.02 & 23.88 & \textbf{48.8} & 66.14 & 83.83 & \textbf{88.42} & 67.8 & 93.87 & 46.04 & 66.77 & 65.16 \\
\midrule
TPT~\cite{tpt} & 65.41 & 23.1 & 42.93 & 66.36 & 84.63 & 87.22 & \textbf{68.86} & 94.12 & \textbf{46.99} & 68.00 & 64.76\\
MTA (Ours) & 64.98 & 25.32 & 38.71 & 68.05 & 84.95 & 88.22 & 68.26 & 94.13 & 45.59 & 68.11 & 64.63 \\
\rowcolor{LightGray} MTA +E. (Ours) & \textbf{66.67} & \textbf{25.2}  & 45.36 & \textbf{68.47} & \textbf{85.00} & 88.24 & 68.06 & \textbf{94.21} & 45.9 & \textbf{68.69} & \textbf{65.58} \\
\bottomrule

\end{tabular}}

\end{table*}

\begin{table*}[t!]
\caption{Comparison of augmentation strategies: RandomCrop Vs Diffusion-based. Note that the test set of each dataset is reduced to 1000 samples for computation reasons and batch size of 128 was used as in the DiffTPT paper~\cite{difftpt} (64 randomly cropped and 63 diffusion-generated images for Diffusion). Hence reported performance can vary from Table \ref{tab:zero_shot_imagenet}. We highlight the best result by {\bf bolding}.}
\label{tab:augs}
\centering
\resizebox{0.7\linewidth}{!}{%
\begin{tabular}{llcccccccc}
\toprule
\multicolumn{1}{l}{\bf Augmentation} 
&\multicolumn{1}{l}{\bf Method}  
&\multicolumn{1}{c}{\makecell{\textbf{ImageNet}}}
&\multicolumn{1}{c}{\makecell{\textbf{-A}}} 
&\multicolumn{1}{c}{\makecell{\textbf{-V2}}} 
&\multicolumn{1}{c}{\makecell{\textbf{R}}}
&\multicolumn{1}{c}{\makecell{\textbf{-Sketch}}}
&\multicolumn{1}{c}{\bf Average} 
\\ \midrule

\multirow{2}{*}{\makecell{RandomCrop}} & TPT~\cite{tpt} 
      & 68.15 & 51.23 & \textbf{66.17} & 76.88 & 49.31 & 62.35 \\
& \cellcolor{LightGray} MTA  & \cellcolor{LightGray} \textbf{69.11} & \cellcolor{LightGray} \textbf{55.27} & \cellcolor{LightGray} 65.71 & \cellcolor{LightGray} \textbf{77.48} & \cellcolor{LightGray} \textbf{50.23} & \cellcolor{LightGray} \textbf{63.56} \\
\midrule
\multirow{2}{*}{\makecell{Diffusion}} 
& DiffTPT~\cite{difftpt}  & 67.83 & 53.43  & \textbf{65.18}  & \textbf{76.85} & 50.2 &    62.7  \\
& \cellcolor{LightGray} MTA & \cellcolor{LightGray} \textbf{69.18} & \cellcolor{LightGray} \textbf{54.5} & \cellcolor{LightGray} 64.81 & \cellcolor{LightGray} 76.82 & \cellcolor{LightGray} \textbf{51.09} & \cellcolor{LightGray} \textbf{63.28}     \\
\bottomrule
\end{tabular}%
}
\end{table*}

\paragraph{Interpretation}
Eqs. \eqref{final-updates-y} and \eqref{fixed-point-iteration-updates} can be nicely interpreted in Figure \ref{fig:interpretation}. Sub-figure \ref{fig:interpretation_inlier} shows how the {\em inlierness} scores are spreading. A data point is given a high {\em inlierness} score if it is close to the mode and/or close to other data points with high {\em inlierness} scores. Therefore, {\em inlierness} scores are spreading iteratively from data points close to the mode toward other data points controlled by their affinity relations. Sub-figure \ref{fig:interpretation_mode} shows the update direction followed by traditional MeanShift, i.e., updates \eqref{fixed-point-iteration-updates} but with fixed $y_p = 1, \, \forall p$. The orange arrows follow the update of our robust MeanShift, i.e. joint updates in \eqref{final-updates-y} and \eqref{fixed-point-iteration-updates}. Our mode update is directed toward dense regions with high {\em inlierness} scores, thus avoiding close regions with few or isolated data points.

\section{Experimental settings}

\subsection{Datasets}
To evaluate our proposed MTA, we follow the setting of previous works~\cite{coop, tpt}. We assess our method on ImageNet~\cite{imagenet} and its four variants (ImageNet-A~\cite{hendrycks2021nae}, ImageNet-V2~\cite{recht2019imagenet}, ImageNet-R~\cite{hendrycks2021many}, ImageNet-Sketch~\cite{wang2019learning}) to measure robustness to natural domain shifts. Additionally, we also consider 10 datasets for fine-grained classification of scenes (SUN397~\cite{sun397}), aircraft types (Aircraft~\cite{aircraft}), satellite imagery (EuroSAT~\cite{eurosat}), automobiles (StanfordCars~\cite{cars}), food items (Food101~\cite{food}), pet breeds (OxfordPets~\cite{pets}), flowers (Flower102~\cite{flower}), general objects (Caltech101~\cite{caltech101}), textures (DTD~\cite{dtd}) and human actions (UCF101~\cite{ucf101}). These diverse datasets provide a comprehensive benchmark for visual classification tasks.
\subsection{Implementation details}
\paragraph{No hyperparameter tuning.} In the MeanShift algorithm, the bandwidth is a sensitive hyperparameter that can cause the algorithm to become stuck in small, locally dense areas if set too low, or escape significant dense regions if set too high~\cite{variable_bandwidth}. We utilize the Gaussian kernel~\cite{CarreiraPerpin2007GaussianMI} as described in Section \ref{sec:method} and adopt a variable bandwidth~\cite{variable_bandwidth} in which each point is assigned a unique bandwidth value $h_{p}^2$. The bandwidth of a point is estimated with a ratio $\rho$ of its closest neighbors: $h_{p}^2=\frac{1}{\rho (N-1)} \sum_{q \in I_p} \|{\mathbf f}_p - {\mathbf f}_q \|^2$ with $\rho$ set to 0.3 inspired by~\cite{scikit-learn}. Here, $I_p$ represents the set of indices corresponding to the neighbors of point p. Initial guess of the mode can also impact the final solution and is set to the embedding of the original (i.e., non-augmented) image. Affinities are based on the text prediction as described in Section \ref{sec:method}. Finally, $\lambda$ and $\lambda_{{\mathbf y}}$ are set to 4 and 0.2 respectively and remain fixed for every CLIP visual encoder and dataset. This leads to a hyperparameter-free method across all experiments. Unless otherwise mentioned, we report top-1 accuracy using the ViT-B/16 backbone.
\paragraph{Comparison methods.} We employ the term ensemble for the majority vote among the 80 predefined handcrafted prompts of CLIP \cite{clip}. For the zero-shot scenario, TPT is performed with one step as suggested in their work~\cite{tpt}. We evaluate the zero-shot setting~\cite{lampert2009learning} with the basic "\texttt{a photo of a} [$\texttt{class}_k$]" as prompt initialization and with the ensemble for final prediction if mentioned. Note that combining ensemble with TPT is not straightforward as the goal is to use the optimized prompt for prediction, hence we only use ensemble with our approach. The weights of CoOp~\cite{coop} are from 16 shots Imagenet with 4 tokens. For the few-shot scenario, the number of tokens of CoOp~\cite{coop} and ProGrad \cite{prograd} are specified in the caption of each table. As proposed in~\cite{multimodality}, to establish a fair comparison between few-shot methods, we limit the number of validation samples to $min(n,4)$ where $n$ is the number of training shots, notably for Tip-Adapter and Tip-Adapter-F~\cite{tip-adapter} that tune hyperparameters. Because of the randomness inherent in test-time augmentation or some training procedures (e.g., prompt tuning), each reported performance is the averaged top-1 accuracy of three different random seeds. More detailed performances are available in Appendix \ref{appendix:additional_results}.
\begin{figure}[t]
    \centering
    \includegraphics[width=\linewidth]{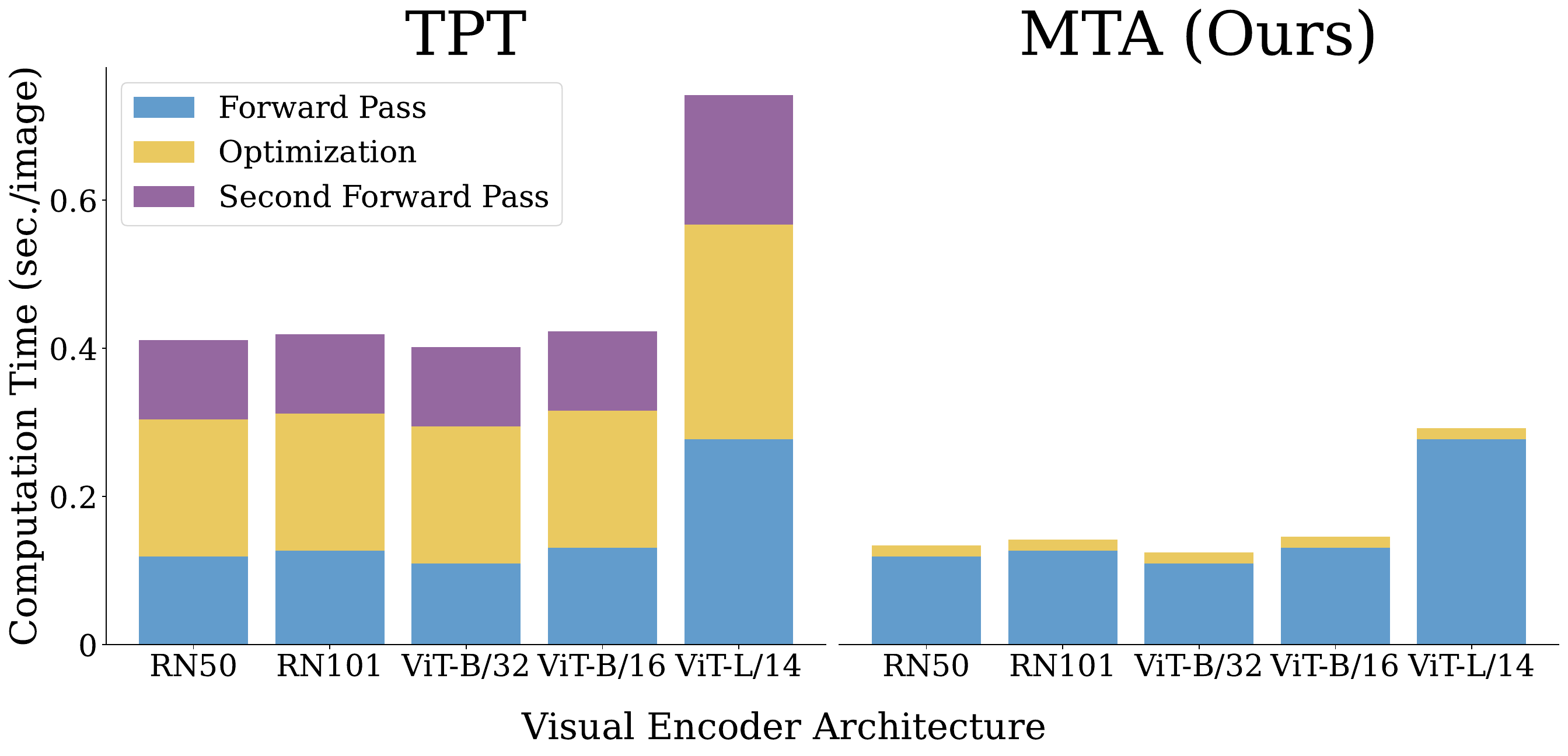}
    \caption{Runtime in seconds per image on ImageNet for TPT and MTA with 5 different backbones: RN50 (ResNet-50), RN101 (ResNet-101), ViT-B/32, ViT-B/16 and ViT-L/14. Experiences were performed on a single A100 40Gb GPU.}
    \label{fig:runtime}
\end{figure}
\paragraph{Test-time augmentation.}
Our proposed method uses random cropping (RandomCrop) as augmented view generator identical to the one in TPT~\cite{tpt}. We also study at the end of Section \ref{sec:zero_shot} the impact of a more complex data augmentation based on diffusion as done in DiffTPT~\cite{difftpt}. Note that TPT and DiffTPT are using slightly stronger augmentations for the 10 fine-grained classification datasets inspired by AugMix~\cite{augmix}. In our case, for a more realistic zero-shot setting, we keep the simple RandomCrop for all the 15 datasets.

\section{Zero-shot}
\label{sec:zero_shot}

\paragraph{MTA globally outperforms TPT while respecting the black-box constraints and running nearly three times as fast.} Table \ref{tab:zero_shot_imagenet} demonstrates MTA's stable improvement over TPT on ImageNet and its variants with both the basic "\texttt{a photo of a} [$\texttt{class}_k$]" and CoOp's pretrained prompt. Moreover, Table \ref{tab:zero_shot_datasets} indicates that, except for the EuroSAT dataset~\cite{eurosat}, MTA consistently enhances baseline performance across fine-grained datasets. It also surpasses TPT when combined with ensembling, which respects the black-box assumption. TPT is not able to outperform the majority vote strategy in average for the 10 fine-grained classification datasets, questioning its applicability for a larger variety of tasks. Finally, we compare runtimes on ImageNet in Figure \ref{fig:runtime}. MTA is nearly three times faster than TPT, notably due to the quick optimization step and the removed second forward pass.

\begin{figure*}[t!]
\centering
\subfloat{\label{sfig:a}\includegraphics[width=.233\textwidth]{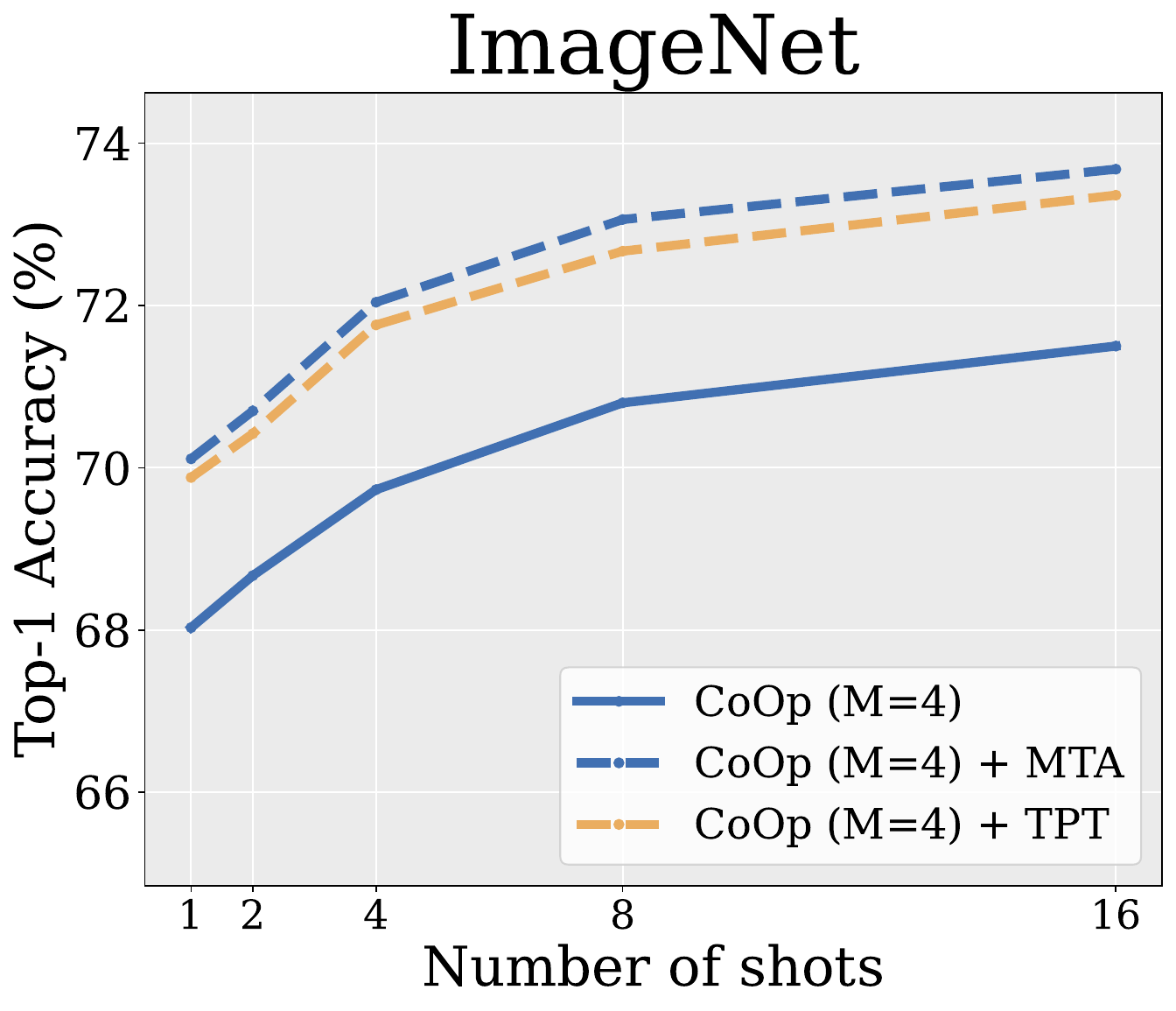}}\hfill
\subfloat{\label{sfig:b}\includegraphics[width=.233\textwidth]{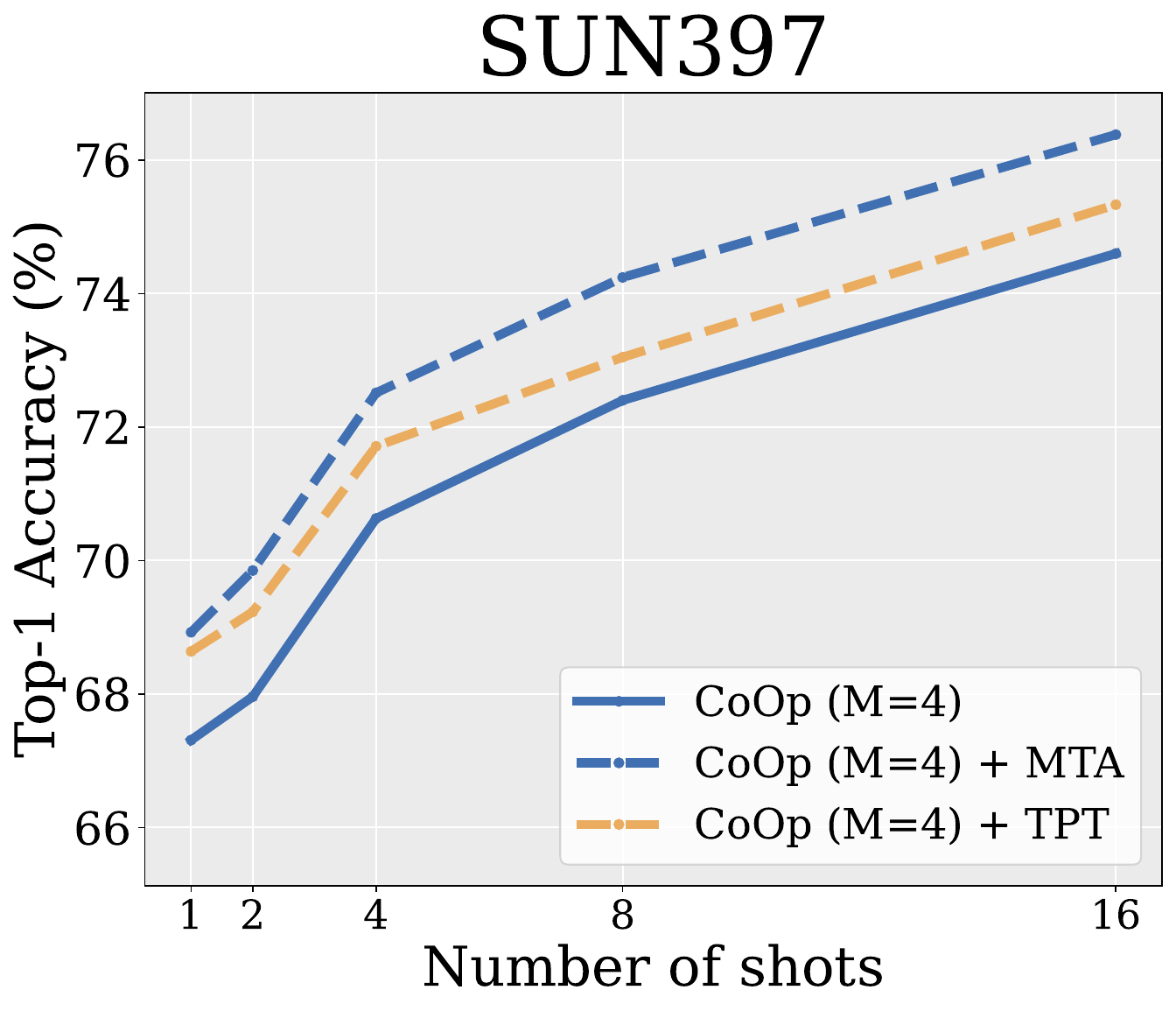}}\hfill
\subfloat{\label{sfig:c}\includegraphics[width=.233\textwidth]{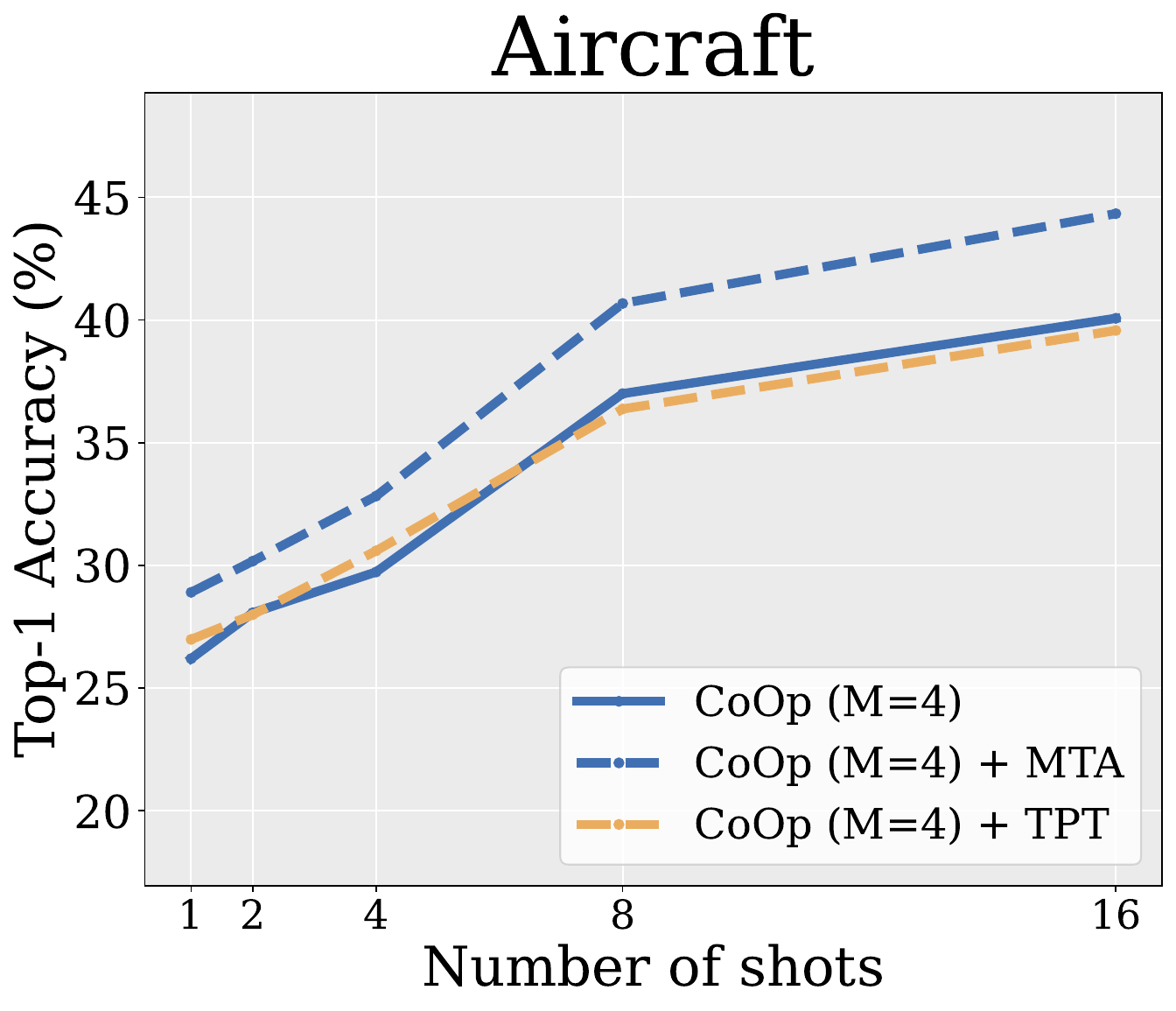}}\hfill
\subfloat{\label{sfig:d}\includegraphics[width=.233\textwidth]{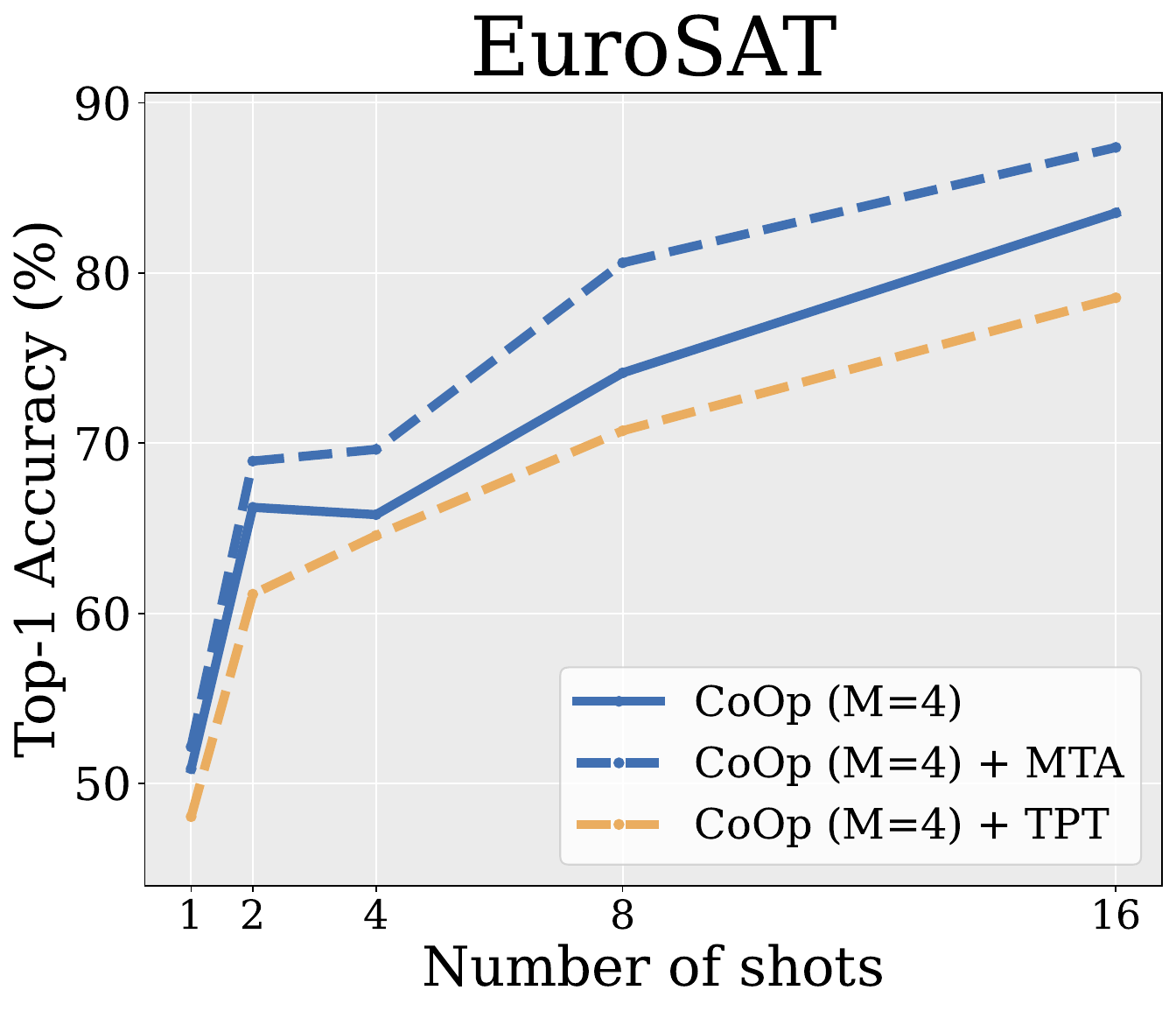}}\hfill
\subfloat{\label{sfig:f}\includegraphics[width=.233\textwidth]{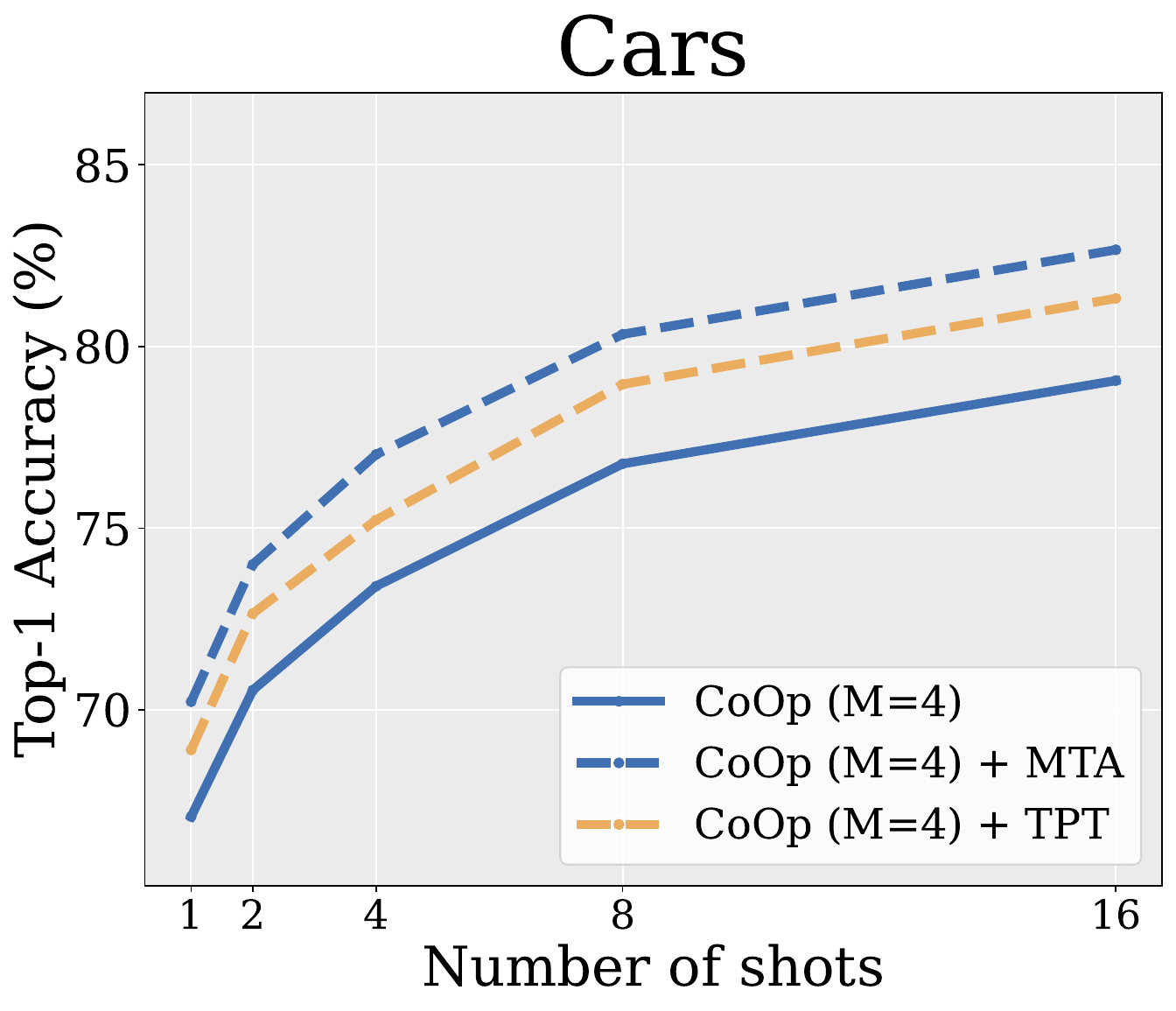}}\hfill
\subfloat{\label{sfig:g}\includegraphics[width=.233\textwidth]{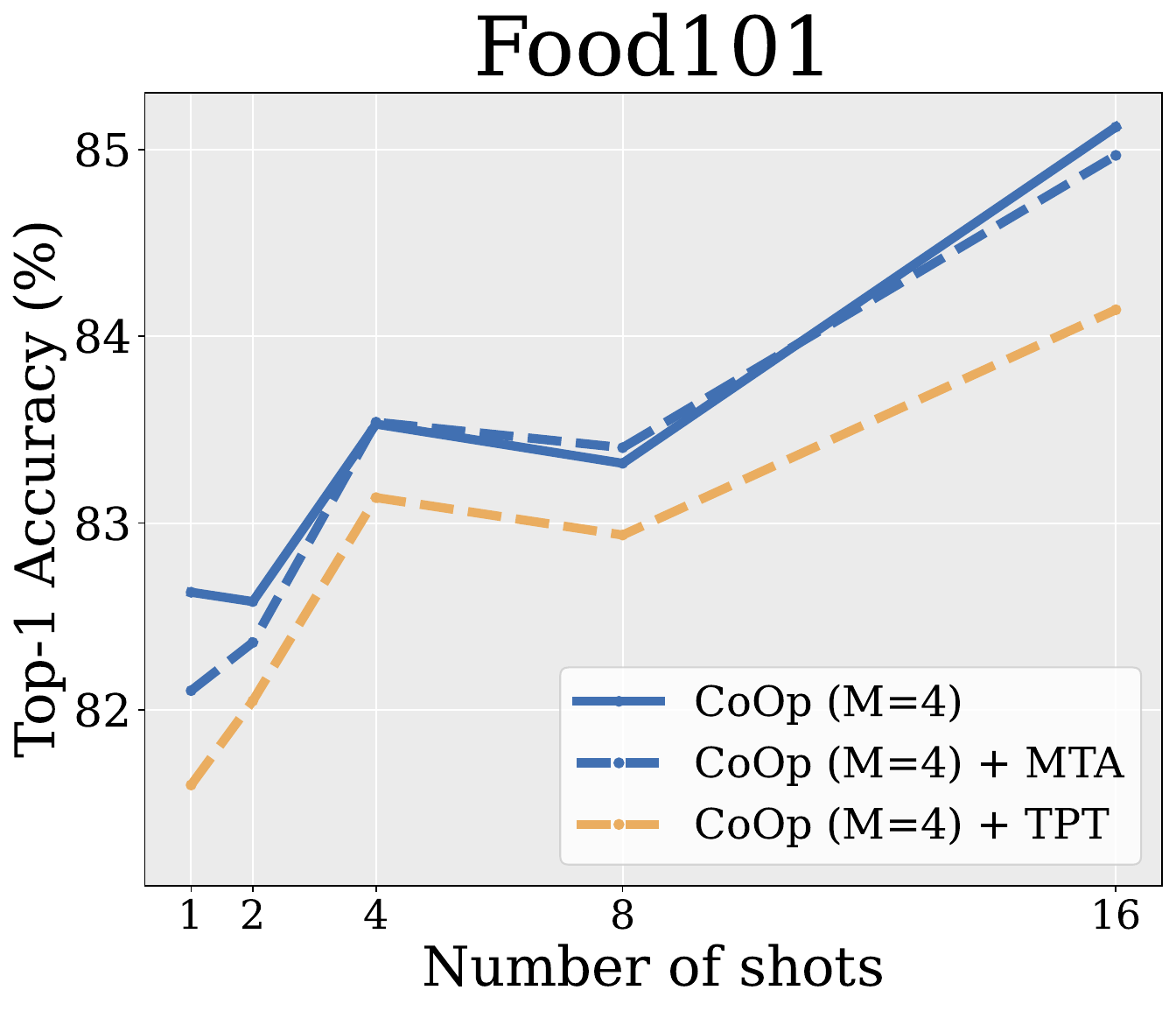}}\hfill
\subfloat{\label{sfig:h}\includegraphics[width=.233\textwidth]{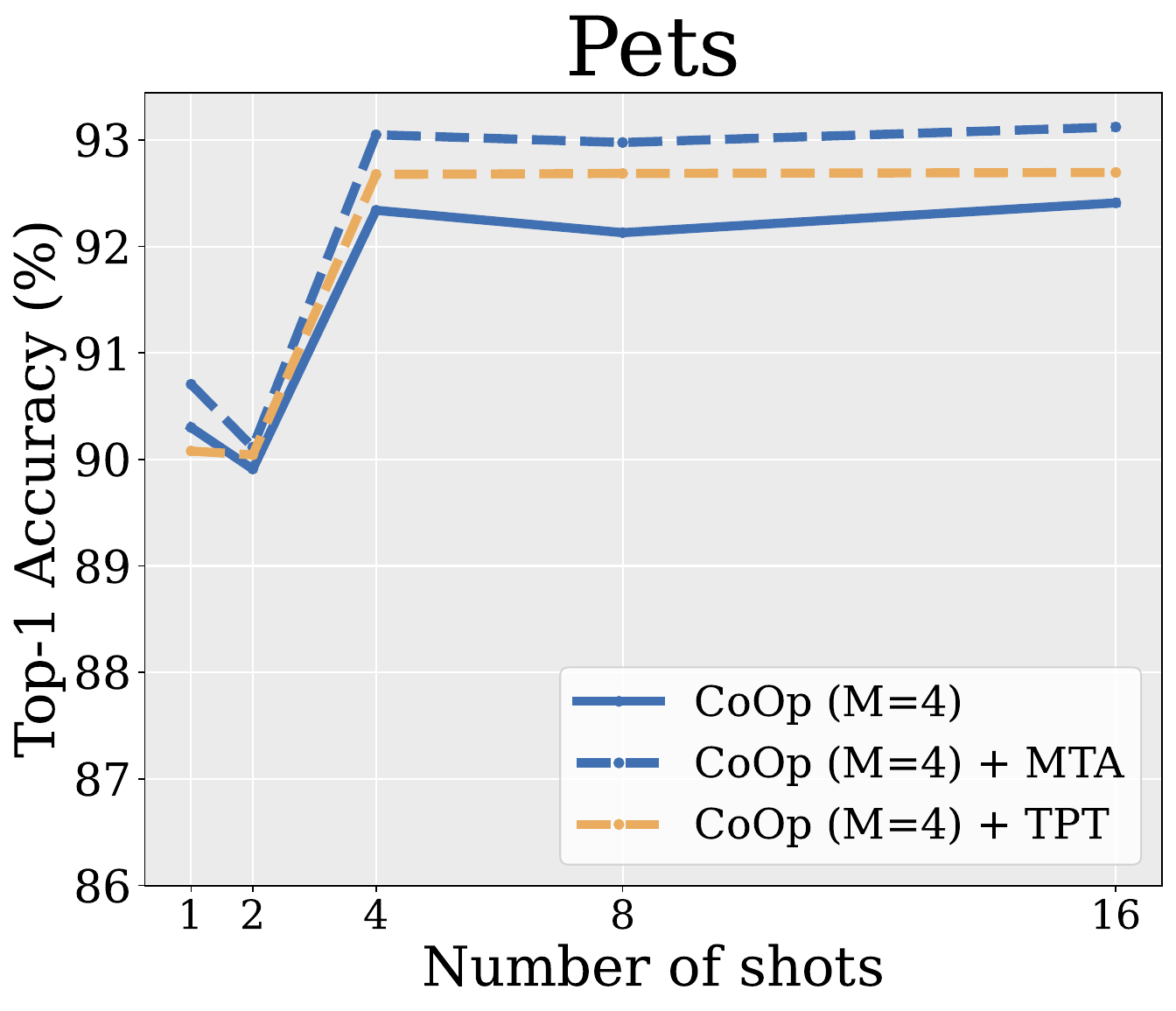}}\hfill
\subfloat{\label{sfig:i}\includegraphics[width=.233\textwidth]{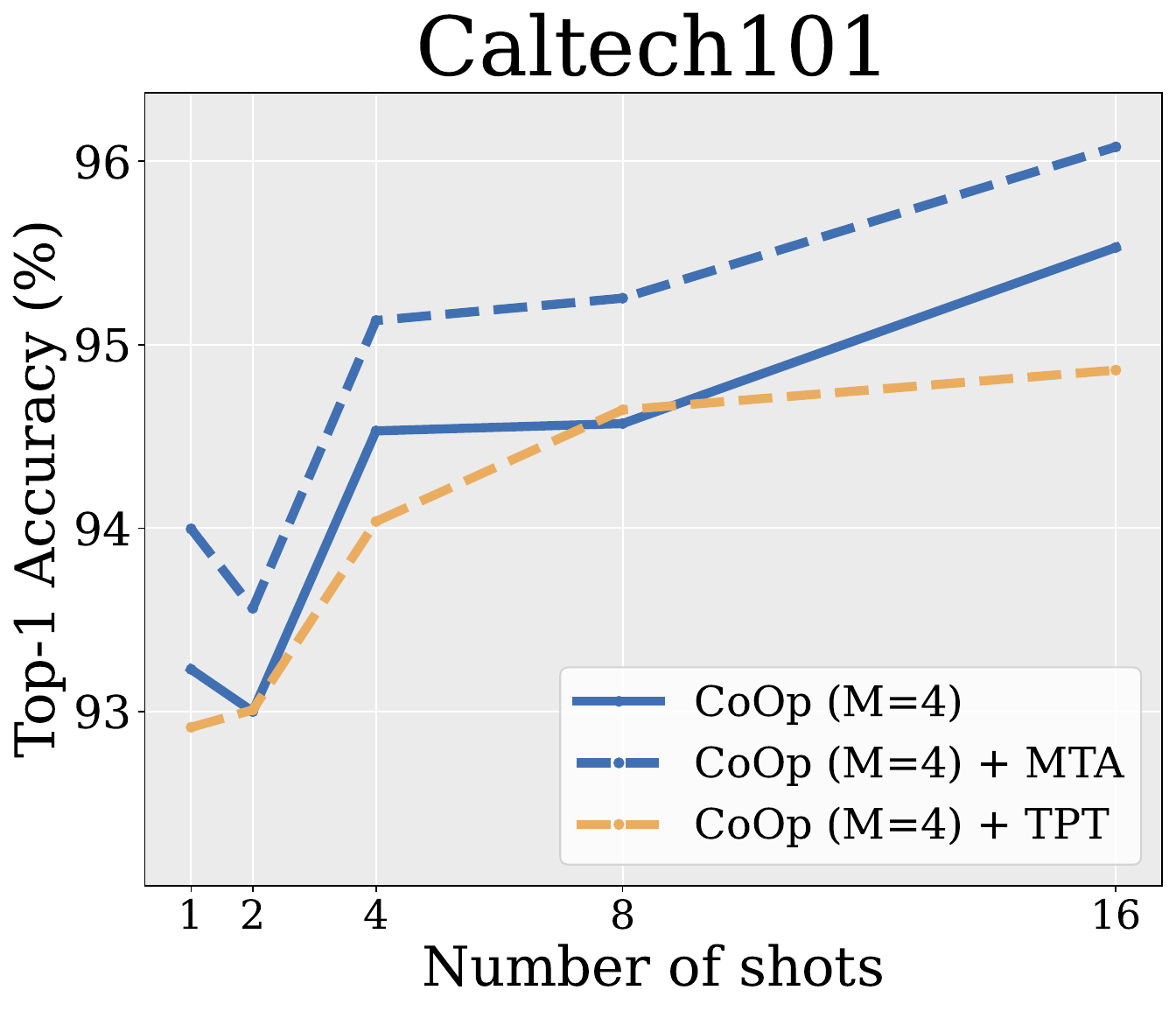}}\hfill
\subfloat{\label{sfig:f}\includegraphics[width=.233\textwidth]{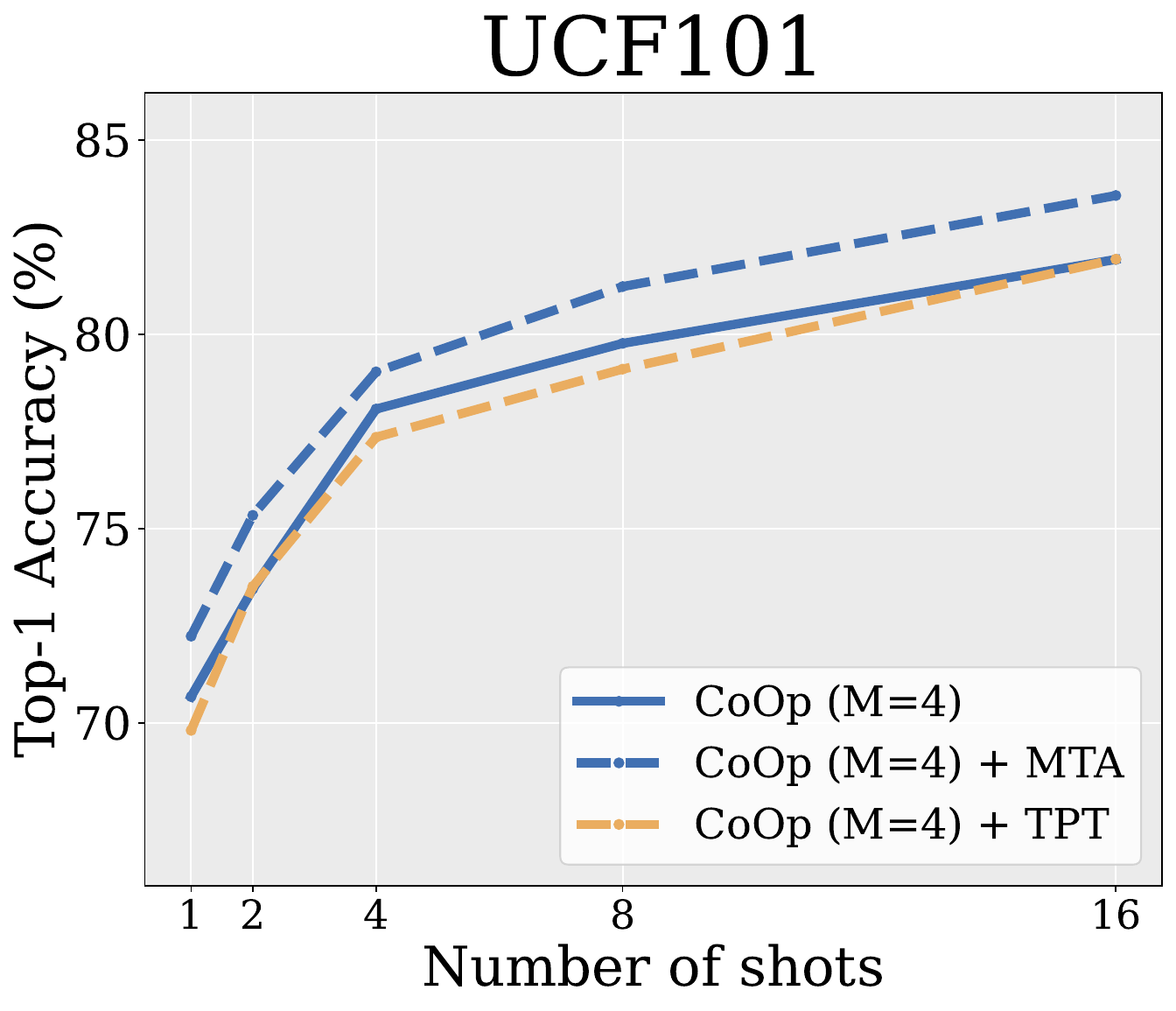}}\hfill
\subfloat{\label{sfig:g}\includegraphics[width=.233\textwidth]{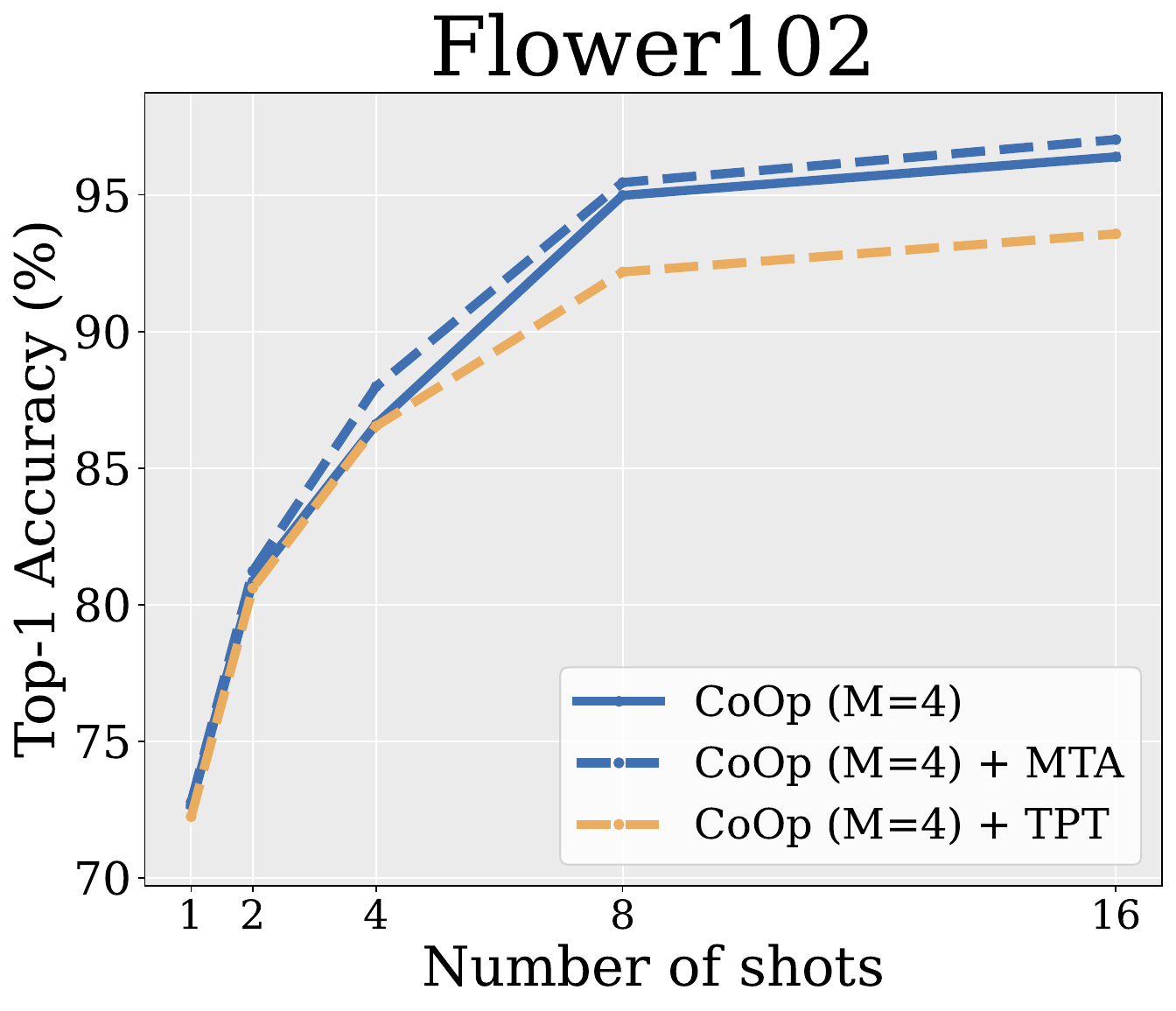}}\hfill
\subfloat{\label{sfig:h}\includegraphics[width=.233\textwidth]{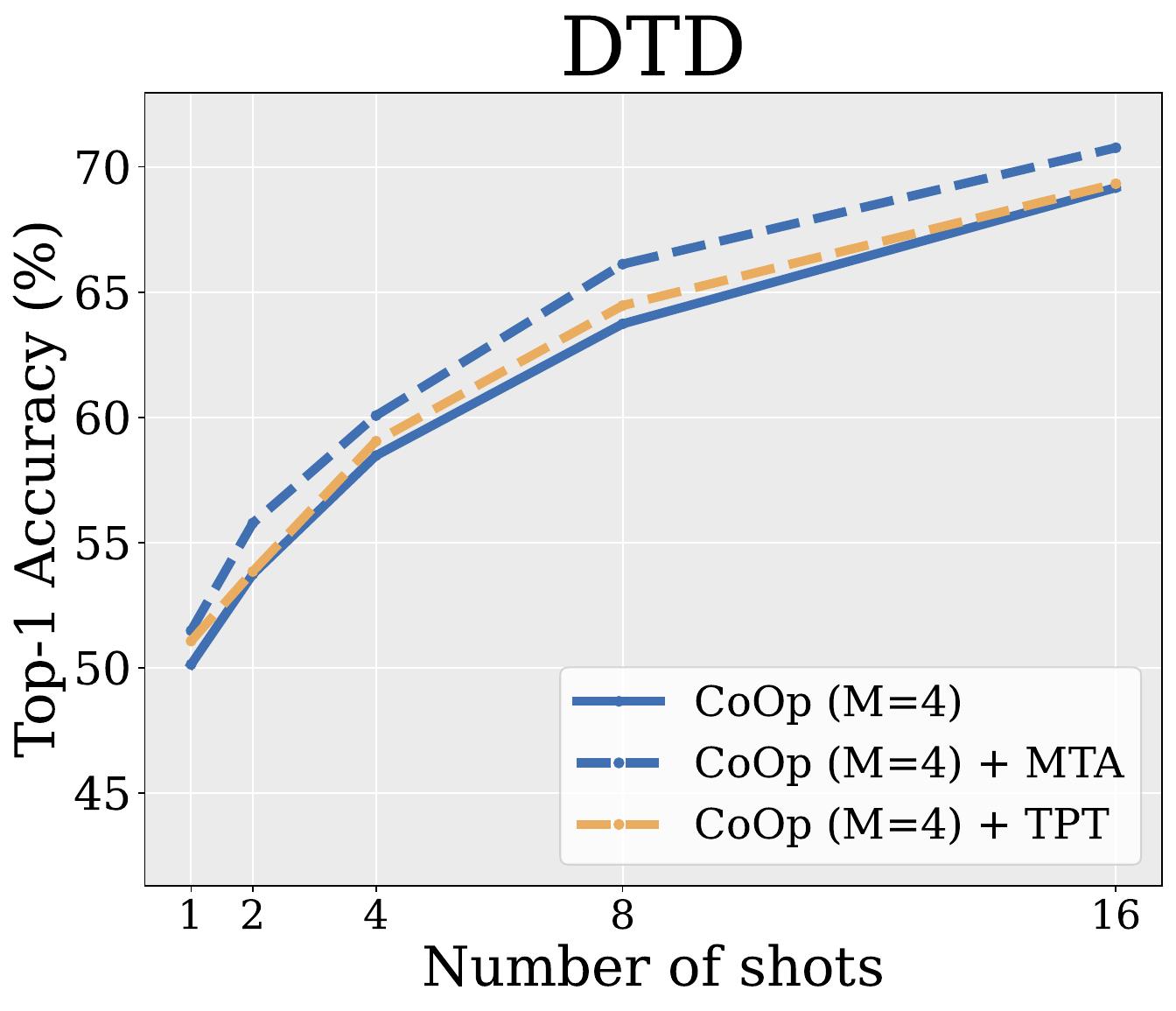}}\hfill
\subfloat{\label{sfig:i}\includegraphics[width=.233\textwidth]{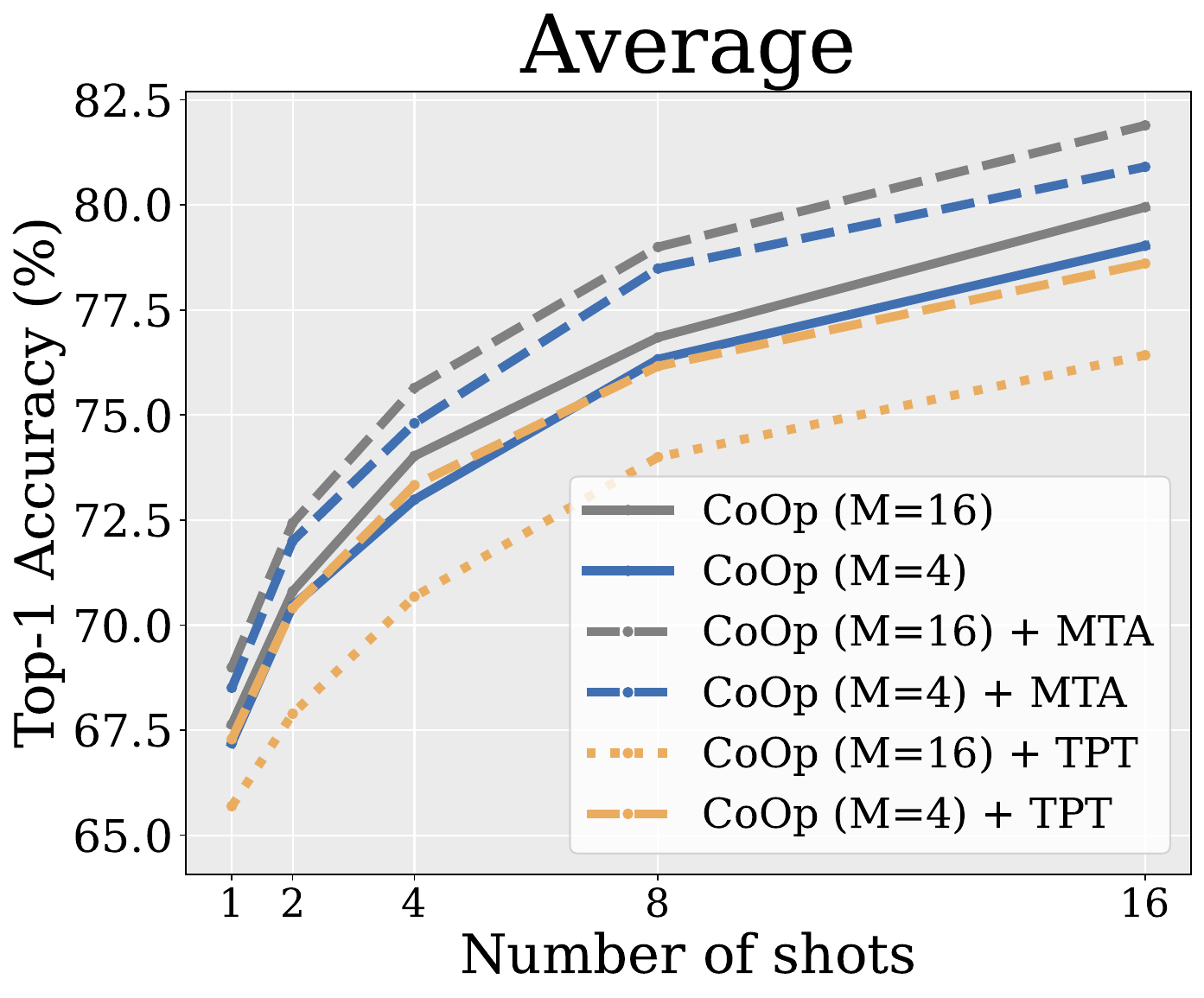}}\hfill
\caption{Few-shot learning results on the 10 fine-grained datasets and ImageNet. We compare MTA and TPT when added on top of CoOp prompts (M=4 tokens) for increasing number of shots. Averaged top-1 accuracy over the 11 datasets is shown on the bottom right, we additionally show the averaged top-1 accuracy for CoOp with M=16 tokens.}
\label{fig:coop_few_shots}
\end{figure*}

\paragraph{MTA benefits from improved prompt strategy.} Table \ref{tab:zero_shot_imagenet} and \ref{tab:zero_shot_datasets} concur in suggesting that the improvements brought by our approach complement those from refined prompt strategies, i.e., basic ensemble of handcrafted prompts or soft pretrained prompts. This may imply that leveraging both modalities in the same optimization process, as suggested in~\cite{multimodality} for the few-shot setting, could further enhance performance.
\begin{table}[t!]
\renewcommand{\arraystretch}{1.1}
\caption{Averaged top-1 accuracy on ImageNet and its 4 variants for different visual encoders of CLIP. Hyperparameters are kept identical across datasets and visual encoder architectures. +E. stands for Ensemble.}
\label{tab:backbones}
\begin{center}
\resizebox{\linewidth}{!}{
\begin{tabular}{lccccc}
\toprule
Architecture & ResNet-50 & ResNet-101 & ViT-B/32 & ViT-B/16 & ViT-L/14
\\ \midrule
CLIP & 44.11 & 49.48 & 50.65 & 59.11 & 70.65 \\
Ensemble
& 46.23 & 51.58 & 52.23 & 61.12 & 72.6 \\
TPT & 47.23 & 52.16 & 53.27 & 62.40 & 73.67 \\
 MTA & 47.22 & 53.15 & 55.17 & 63.16 & 73.88 \\
\rowcolor{LightGray}
MTA + E. & \textbf{48.35} & \textbf{54.23} & \textbf{56.1} & \textbf{64.06} & \textbf{74.71} \\
\bottomrule
\end{tabular}}
\end{center}
\end{table}
\paragraph{MTA bridges the transferability gap of pretrained prompts due to better image representation.} We use CoOp pretrained on ImageNet to measure the domain generalization ability of our method as in~\cite{coop} in Table \ref{tab:zero_shot_imagenet}. Although the gains of CoOp over the ensembling strategy remain ambiguous, the combination of CoOp with MTA significantly enhances accuracy for all datasets. This notable improvement suggests that the mode found by MTA is able to highlight the generalization ability of pretrained prompts. This illustrates an additional use case of MTA, which can be used in synergy with fine-tuning for a specific task (e.g., prompt tuning). This aspect is emphasized in Section \ref{sec:few_shot}.

\paragraph{MTA generalizes across visual encoder architectures with the same hyperparameters.} Generalization across various model architectures is a desirable attribute of zero-shot methods, as it eliminates the need for laborious and computationally demanding hyperparameter tuning. This becomes increasingly critical in the context of the current scaling trend, where model complexity is rapidly expanding~\cite{brown2020language, chowdhery2022palm}. Table \ref{tab:backbones} shows consistent improvement on the baseline for five different visual backbones of CLIP with fixed hyperparameters. It demonstrates the generalization ability of our method across architectures and model scales.

\paragraph{MTA is applicable to other data augmentation strategies.} We explore the compatibility of MTA with different types of data augmentation. Specifically, we follow the protocol of DiffTPT~\cite{difftpt} to generate augmented views from cropping and diffusion model~\cite{stable_diffusion}. We employ a batch of 128 images composed of the original one, 63 diffusion-generated and 64 randomly cropped views. Since generating images by diffusion is more computationally intensive than RandomCrop (generating 63 images takes approximately 2 minutes on a A100 40Gb GPU), we present these results separately. Inspired by the observations of DiffTPT about the reliance of the images generated by diffusion, we increase $\rho$ to a slightly less restrictive value of 0.5 for the purpose of this experiment. Otherwise, we keep the same hyperparameters $\lambda$ and $\lambda_{{\mathbf y}}$ and do not treat the two kinds of augmentation differently. As demonstrated in Table \ref{tab:augs}, our method surpasses DiffTPT on average, further evidencing its effectiveness across a broad range of applications.

\section{Few-shot}
\label{sec:few_shot}

\paragraph{Fine-grained learned prompts benefit from MTA but not from TPT.} As depicted in Figure \ref{fig:coop_few_shots}, MTA improves CoOp performances across shots and datasets with notably Aircraft 40.07\% $\rightarrow{}$ 44.33\% (\textbf{+ 4.26\%}), EuroSAT  83.53\% $\rightarrow{}$ 87.38\% (\textbf{+ 3.85\%}), and  Cars 79.06\% $\rightarrow{}$ 82.66\% (\textbf{+ 3.6\%}) with 16 shots. In contrast, TPT generally diminishes performance across most datasets,  highlighting its limitations in enhancing prompts for fine-grained tasks. While the accuracy for 16 tokens is on average higher for MTA and lower for TPT, we report the 4 tokens results in alignment with the TPT paper~\cite{tpt} to maintain fairness. Detailed performances for 16 tokens and individual datasets are available in Appendix \ref{appendix:additional_results}.
\paragraph{MTA can be applied atop few-shot learning methods.} As demonstrated in Table \ref{tab:few_shot_methods}, applying MTA to prompt-based and adapter-based few-shot learning methods on ImageNet results in notable performance improvements. Specifically, prompt-based methods see an average gain exceeding \textbf{2\%} across shots, with adapter-based methods showing slightly lower yet significant improvements. Indeed, our affinity term, outlined in Eq. \ref{eq:affinity_term}, can greatly benefit from refined prompts. Nevertheless, we can notice that a training-free approach, Tip-Adapter, combined with MTA is competitive with prompt tuning methods without MTA. This could present a compelling trade-off: opting for more intensive computational efforts during training or at test-time.%opting for more intensive computational efforts during training (as in prompt tuning) or at test-time (as with MTA).
\begin{table}[t]
\renewcommand{\arraystretch}{1}
\caption{Improvement of few-shot learning methods on ImageNet when MTA is added on top. CoOp and ProGrad are using 16 tokens. $\Delta$ highlights the gain. \cmark \hspace{0.1cm} for training-free methods, \xmark \hspace{0.1cm} otherwise.}
\label{tab:few_shot_methods}
\begin{center}
\resizebox{\linewidth}{!}{
\begin{tabular}{lcccccc}
\toprule
\multicolumn{1}{l}{Shots}
&\multicolumn{1}{c}{}
&\multicolumn{1}{c}{1} 
&\multicolumn{1}{c}{2}
&\multicolumn{1}{c}{4}
&\multicolumn{1}{c}{8} 
&\multicolumn{1}{c}{16}
\\
\midrule
Tip-Adapter~\cite{tip-adapter}    &    \cmark    & 68.94 & 69.18 & 69.75 & 70.15 & 70.51 \\
Tip-Adapter-F~\cite{tip-adapter}    &    \xmark    & 69.36 & 69.95 & 70.74 & 71.82 & 73.39 \\
CoOp~\cite{coop} & \xmark  & 65.7 & 66.97 & 68.83 & 70.57 & 71.87 \\
ProGrad~\cite{prograd} &  \xmark   & 67.01 & 69.06 & 70.15 & 71.25 & 72.14 \\
\hline \hline 
Tip-Adapter + MTA & \cmark   & 71.05 & 71.12 & 71.4 & 71.9 & 72.05\\
$\Delta$ & & (\textcolor{darkerGreen}{+2.11}) & (\textcolor{darkerGreen}{+1.94}) & (\textcolor{darkerGreen}{+1.65}) & (\textcolor{darkerGreen}{+1.75}) & (\textcolor{darkerGreen}{+1.54})\\

Tip-Adapter-F + MTA & \xmark   & 71.35 & 71.48 & 72.17 & 73.18 & 74.23 \\
$\Delta$ & & (\textcolor{darkerGreen}{+1.99}) &  (\textcolor{darkerGreen}{+1.53}) & (\textcolor{darkerGreen}{+1.43}) & (\textcolor{darkerGreen}{+1.36}) & (\textcolor{darkerGreen}{+0.84}) \\
CoOp + MTA & \xmark   & 67.82 & 69.1 & 71.05 & 72.85 & 74.20\\
$\Delta$ & & (\textcolor{darkerGreen}{+2.12}) & (\textcolor{darkerGreen}{+2.13}) & (\textcolor{darkerGreen}{+2.22}) & (\textcolor{darkerGreen}{+2.28}) & (\textcolor{darkerGreen}{+2.33}) \\
ProGrad + MTA &  \xmark  & 69.27 & 71.39 & 72.50 & 73.66 & 74.41\\
$\Delta$ & & (\textcolor{darkerGreen}{+2.26}) & (\textcolor{darkerGreen}{+2.33}) & (\textcolor{darkerGreen}{+2.35}) & (\textcolor{darkerGreen}{+2.41}) & (\textcolor{darkerGreen}{+2.27}) \\

\bottomrule
\end{tabular}}
\end{center}
\end{table}

\section{Ablation study}
\label{sec:ablation}
% We perform ablative analysis on various components of MTA, namely the number of augmented views, the {\em inlierness} score and the affinity measure of Equation \ref{}.

\begin{figure}[t]
    \centering
    \includegraphics[width=\linewidth]{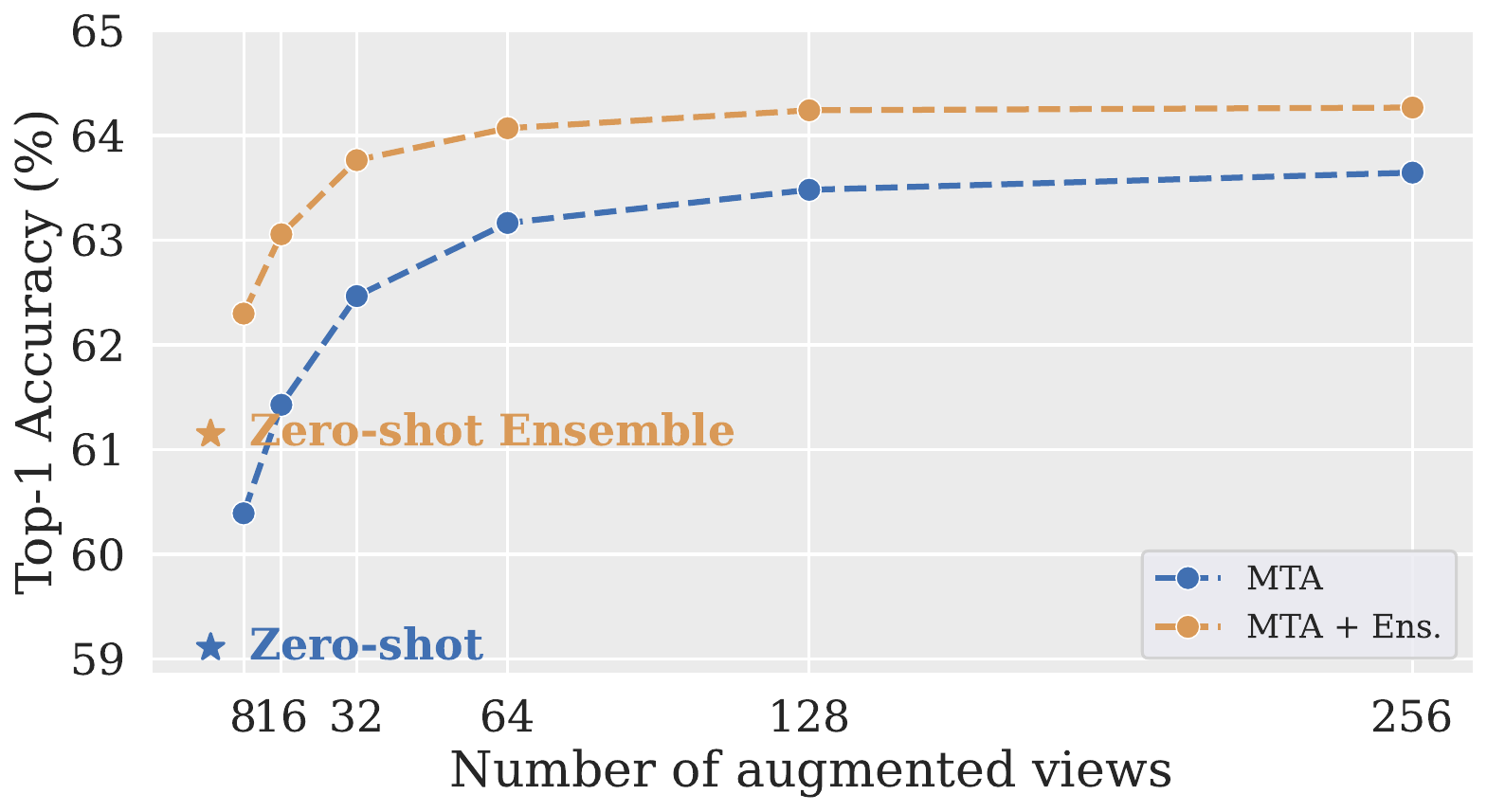}
    \caption{Averaged top-1 accuracy of MTA with and without majority vote for final prediction on the 5 ImageNet variants with increasing number of augmented views.}
    \label{fig:mean_shift_sizes}
\end{figure}

\paragraph{Number of augmented views.}
In Figure \ref{fig:mean_shift_sizes}, the accuracy for MTA and MTA + Ensemble increases as the number of augmented views grows until reaching a plateau around 128. Even when we restrict the number of augmented views to 16, our method still brings about \textbf{2.3}\% gain, which makes it a useful tool for lighter applications, up until \textbf{4.4}\% gain for batches of 128 views. We can observe similar gains when combined with ensemble of prompts, with \textbf{1.9}\% gain for batches of 16 up until \textbf{3.1}\% for larger batches.

\paragraph{{\em Inlierness} scores.} We compare performance with equal weights on each augmented view (i.e., traditionnal MeanShift), with a confidence threshold as in TPT~\cite{tpt} and with our {\em inlierness} formulation in Table \ref{tab:inlier_ablation}. The {\em inlierness} formulation yields better performance on average over the 15 datasets. Note that the relatively high score of confidence threshold is mainly due to peak performance on ImageNet-A, a trend not consistent on other datasets, see Appendix \ref{appendix:additional_results}. Additionally, Appendix \ref{appendix:further_details} contains a study of $\lambda$ and $\lambda_{{\mathbf y}}$ showing their interdependent relation and importance.
% \paragraph{Mode versus Mean.} Again, note that the relatively high score is mainly due to high performance on ImageNet-A ($60 \%$) which is not observed on other datasets. Nonetheless a simple average of the selected embeddings are competitive with TPT, requestionning the interest of prompt tuning in its current state. 
\paragraph{Affinity measure.} Rather than using the affinity measure based on text described in Section \ref{sec:method}, one could use only visual features $({\mathbf f}_p)_{1 \leq p \leq N}$ to compute the affinity between two embeddings $w_{p,q} = {\mathbf f}_p^t {\mathbf f}_q$. For the sake of fairness, we reperform a comprehensive hyperparameter search ($\lambda$ ranging from 0.1 to 10 and $\lambda_{{\mathbf y}}$ from 0.1 to 0.5). Best performance is reported in Table \ref{tab:inlier_ablation}. We observe that vision-based affinity measure degrades performance in comparison to text-based affinity measure.

\begin{table}[h]
\renewcommand{\arraystretch}{1}
\caption{Ablation study on two main components of MTA: the {\em inlierness} scores and the affinity measure. Reported value is the averaged top-1 accuracy over the 15 datasets studied in this work.}\label{tab:inlier_ablation}
    \centering
    \resizebox{0.8\linewidth}{!}{
    \begin{tabular}{llc}
    \toprule
      Baseline & CLIP & 62.09 \\
    \midrule 
     \multirow{3}{*}{\makecell{Filtering \\ Strategy}} & MeanShift (no {\em inlierness} scores) & 59.93 \\
     & Confidence thresh. ($10\%$) & 63.27 \\
     & \cellcolor{LightGray} {\em Inlierness} scores & \cellcolor{LightGray} \textbf{64.14} \\
    \midrule 
     \multirow{2}{*}{\makecell{Affinity \\ Measure}} & Vision-based & 63.15 \\
     & \cellcolor{LightGray} Text-based & \cellcolor{LightGray} \textbf{64.14} \\
     \bottomrule
    \end{tabular}}
\end{table}

\section{Conclusion}
\label{sec:conclusion}
In this work, we have investigated a novel approach to handle test-time augmentation for vision-language models. Our \textbf{M}eanShift for \textbf{T}est-time \textbf{A}ugmentation (MTA) is based on a robust generalization of a well-known mode seeking algorithm, and operates solely on the final embeddings, in a training-free manner. Extensive experiments demonstrate that our method not only surpasses test-time prompt tuning alternatives but also runs significantly faster. Without any other requirements, MTA can easily be deployed in a zero-shot manner and atop few-shot learning methods. We believe our work could serve as a starting point to broaden the current research focus on improving zero-shot vision-language models, and to investigate more efficient approaches, beyond prompt learning.
% In this work, we have investigated a novel approach to handle test-time augmentation for vision-language models. While current literature focuses on prompt tuning, our proposed algorithm, inspired by MeanShift operates solely on the final embedding states of images and prompts in a training-free manner. Our robust \textbf{M}eanShift for \textbf{T}est-time \textbf{A}daptation (MTA) associates an {\em inlierness} score to each augmented view which are optimized concurrently with the mode. Our findings demonstrate that our method not only surpasses test-time prompt tuning alternatives but also operates significantly faster. Without any other requirements, MTA can easily be deployed in a zero-shot manner or atop few-shot learning methods.
\newpage
{
    \small
    \bibliographystyle{ieeenat_fullname}
    \bibliography{mybib}
}

% \input{sec/0_abstract}    
% \input{sec/1_intro}
% \input{sec/2_formatting}
% \input{sec/3_finalcopy}
% {
%     \small
%     \bibliographystyle{ieeenat_fullname}
%     \bibliography{main}
% }

\clearpage
%\setcounter{page}{1}
%\maketitlesupplementary
\appendix

\section{Proof of Equation \ref{final-updates-y}}
\label{appendix:proof_update_y}

%\begin{equation}
%\label{L-bound}
%{\cal L} ({\mathbf y}, {\mathbf m}) \cleq - \sum_{p=1}^N y_p K({\mathbf f}_p - {\mathbf m}) - \lambda \mathbf{y}^t W^t \mathbf{y}^{(n)} - \lambda_{\mathbf{y}} H({\mathbf y})
%\end{equation}
%Solving the Karush-Kuhn-Tucker (KKT) conditions for minimizing bound
%\eqref{L-bound}, s.t. simplex constraint ${\mathbf y} \in \Delta^{K-1}$, gives the following updates for $\mathbf{y}$:
%\begin{align}
%	\label{final-updates-y}
%  y_p^{n+1} = \frac{\exp{ \left (K({\mathbf f}_p - {\mathbf m}) + \lambda \sum_{q = 1}^N w_{p,q} y_q^{(n)}} \right)}{\sum_{j=1}^N \exp{ \left (K({\mathbf f}_j - {\mathbf m}) + \lambda \sum_{q = 1}^N w_{j,q} y_q^{(n)}} \right)}
%\end{align}

%%%%%%%%%%%%%%%%%%%%%%%%%%%%%%%%%%%%%%%%%%%%%%%%%%%%%%

Here, we give details of the derivation of the ${\mathbf y}$-updates in Eq. \eqref{final-updates-y} from the upper bound (majorizing function) in \eqref{L-bound}. Given a solution ${\mathbf y}^{(n)}$ at iteration $n$, the goal is to find the next iterate ${\mathbf y}^{(n+1)}$ that minimizes the following tight upper bound, s.t. simplex constraint ${\mathbf y} \in \Delta^{N-1}$:  
\begin{equation}
\label{L-bound}
{\cal B} (\mathbf{y}, \mathbf{y}^{(n)}) = - \mathbf{y}^t \mathbf{k} 
 - \frac{\lambda}{2} \mathbf{y}^t W^t \mathbf{y}^{(n)} - \lambda_{\mathbf{y}} H({\mathbf y})
\end{equation}
where $\mathbf{k} = (K(\mathbf{f}_p - {\mathbf m}))_{1\leq p \leq N}$.
%Solving the Karush-Kuhn-Tucker (KKT) conditions for minimizing bound
%\eqref{L-bound}, , gives the following updates for $\mathbf{y}$:

The objective function of \eqref{L-bound} is strictly convex. Taking into account the simplex constraint on
${\mathbf y}$, the associated Lagrangian reads:
\begin{equation}
{\cal L} (\mathbf{y}, \mathbf{y}^{(n)}) = - \mathbf{y}^t \mathbf{k} - \frac{\lambda}{2} \mathbf{y}^t W^t \mathbf{y}^{(n)} - \lambda_{\mathbf{y}} H({\mathbf y}) + \gamma (\mathbf{y}^t {\mathbf 1}_N - 1)
\end{equation}
where $\gamma$ is the Lagrange multiplier for simplex constraint ${\mathbf y} \in \Delta^{N-1}$ and ${\mathbf 1}_N$ is the vector of ones. Note that we do not impose explicitly the constraints on the non-negativity of the components of ${\mathbf y}$ because these are implicitly enforced with the entropic barrier term in (\ref{L-bound}), i.e.,  $- \lambda_{\mathbf{y}} H({\mathbf y})$.    
Now, computing the gradient of ${\cal L} (\mathbf{y}, \mathbf{y}^{(n)})$ w.r.t $\mathbf{y}$ yields: 
\begin{equation}
\label{eq:lagrange_gradient}
\nabla_\mathbf{y} {\cal L} (\mathbf{y}, \mathbf{y}^{(n)}) =
- \mathbf{k} - \lambda W^t \mathbf{y}^{(n)} + (\gamma + \lambda_{\mathbf{y}}) {\mathbf 1}_N + \lambda_{\mathbf{y}} \log({\mathbf y})
\end{equation}
By setting the gradients of \eqref{eq:lagrange_gradient} to $0$, we get the optimal solution:
\begin{align}
\label{final-solution-lagrangian}
&y_p^{(n+1)} =\exp \left ((K({\mathbf f}_p - {\mathbf m}) + \lambda \sum_{q = 1}^N w_{p,q} y_q^{(n)}) / \lambda_{\mathbf{y}} \right) . \nonumber \\
&\exp \left (-(\gamma + \lambda_{\mathbf{y}}) \right )
\end{align}
Using this expression in the simplex constraint $\sum_p y_p^{(n+1)} = 1$ enables to recover the following expression of 
$\exp \left (\gamma + \lambda_{\mathbf{y}} \right )$:
\[\sum_{j=1}^N \exp\left ( (K({\mathbf f}_j - {\mathbf m}) + \lambda \sum_{q = 1}^N w_{j,q} y_q^{(n)})/\lambda_{\mathbf{y}} \right) \]
Plugging this expression back in (\ref{final-solution-lagrangian}), we get the final updates:
\begin{align}
	\label{final-updates-y-appendix}
 y_p^{(n+1)} = \frac{\exp \left ((K({\mathbf f}_p - {\mathbf m}) + \lambda \sum_{q = 1}^N w_{p,q} y_q^{(n)})/\lambda_{\mathbf{y}} \right)}{\sum_{j=1}^N \exp \left ((K({\mathbf f}_j - {\mathbf m}) + \lambda \sum_{q = 1}^N w_{j,q} y_q^{(n)})/\lambda_{\mathbf{y}} \right)}
\end{align}

%Cauchy sequence and convergent sequence proof: 

\section{Cauchy and convergent sequence proof}
\label{appendix:cauchy_sequence}
Let us consider iteration $l$, with the associated current {\em inlierness} scores ${\mathbf y}$. Let us prove that $\{ {\mathbf m}^{l}\}_{l \in \mathbb N}$ is a Cauchy sequence. Recall the recursive relation:
\begin{align}
    \mathbf{m}^{l+1} = \frac{\sum_{p = 1}^{N}  y_p K(\mathbf{f}_p - \mathbf{m}^l) \mathbf{f}_p}{\sum_{p = 1}^N y_p  K(\mathbf{f}_p - \mathbf{m}^l)}  
\end{align}
with $K(\mathbf{f}_p - \mathbf{m}^l) = \exp(-\|\frac{\mathbf{f}_p -  \mathbf{m}^{l}}{h}\|^2)$, for some $h > 0$. We define:
\begin{align}
    k(x) &= \exp(-x) \\
    u^l &= \sum_{p = 1}^{N}  y_p K(\mathbf{f}_p - \mathbf{m}^l) \\
    v^l &= \sum_{p = 1}^{N}  y_p K(\mathbf{f}_p - \mathbf{m}^l) \mathbf{f}_p
\end{align}
\textbf{Step 1: } First, let us prove that $\{u^l\}_{l \in \mathbb N}$ is a Cauchy sequence. Recall that in a metric space, a convergent sequence is necessarily a Cauchy sequence. Therefore, we only need to show that $u^l$ is convergent (i.e bounded and strictly monotonic). \\
Notice that for $x>0$, $0 \leq k(x) \leq 1$. Therefore:
\begin{align}
    u^l &= \sum_{p = 1}^{N} y_p k(\| \frac{\mathbf{f}_p -  \mathbf{m}^{l}}{h}\|^2) \\
        &\leq \sum_{p = 1}^{N} y_p \leq 1
\end{align}
Therefore, $u^l$ is bounded between 0 and 1. Now, let us study the consecutive differences $\Delta^l = u^{l+1} - u^{l}$:
\begin{align}
    \Delta^l = \sum_{p = 1}^{N}  y_p \left[k(\frac{\|\mathbf{f}_p -  \mathbf{m}^{l+1}\|^2}{h^2}) - k(\frac{\| \mathbf{f}_p -  \mathbf{m}^{l} \| ^2}{h^2}) \right]
\end{align}
Because $k$  is convex, one can say that $\forall a, b \in \mathbb R$:
\begin{align}
    k(a) - k(b) \geq k'(b)(a-b)
\end{align}
And because $k'(b)=-k(b)$ in our case, one ends up with:
\begin{align}
    k(a) - k(b) \geq k(b)(b-a)
\end{align}
Applied with $a = \frac{\|\mathbf{f}_p -  \mathbf{m}^{l+1} \|^2}{h^2}$ and $b=\frac{\|\mathbf{f}_p -  \mathbf{m}^{l}\|^2}{h^2}$, one can obtain:
\[
\begin{split}
    \Delta^l \geq & \sum_{p = 1}^{N}  y_p K(\mathbf{f}_p - \mathbf{m}^{l}) \left[ \frac{\| \mathbf{f}_p -  \mathbf{m}^{l} \| ^2}{h^2} - \frac{\| \mathbf{f}_p -  \mathbf{m}^{l+1} \| ^2}{h^2} \right] \\ 
             =& \frac{1}{h^2} \sum_{p = 1}^{N} y_p K(\mathbf{f}_p - \mathbf{m}^{l}) \left[ \|\mathbf{m}^{l}\|^2 - \|\mathbf{m}^{l+1}\|^2  \right.\\
               &\left.  -2<\mathbf{m}^{l}, \mathbf{f}_p> + 2 <\mathbf{m}^{l+1}, \mathbf{f}_p> \right] \\
\end{split}
\]
Now is time to recall recursive relation $\mathbf{m}^{l+1} = \displaystyle\frac{v^l}{u^l}$. By simply  expanding, one can end up with:
\begin{align}
    \Delta^l \geq& \frac{1}{h^2}\left[ \| \mathbf{m}^{l}\|^2 u^l - \frac{\|v^l \|^2}{u^l} - 2 <\mathbf{m}^{l}, v^l> + 2 \frac{\| v^l \|^2}{u^l}\right] \nonumber\\
                =& \frac{1}{h^2}u^l \left[ \|\mathbf{m}^{l}\|^2 -2<\mathbf{m}^{l},\mathbf{m}^{l+1}> + \| \mathbf{m}^{l+1} \|^2 \right] \nonumber\\
                \label{eq:relation_mn_un}=& \frac{1}{h^2}u^l \| \mathbf{m}^{l}-\mathbf{m}^{l+1}\| ^2
\end{align}
Therefore, $\Delta^l >0$, which shows that $\{u^l\}_{l \in \mathbb N}$ is strictly increasing. This concludes the proof that $u^l$ is a convergent sequence, and therefore a Cauchy one.\\

\textbf{Step 2: } Now, on top of concluding the proof that $\{u^l\}_{l \in \mathbb N}$ is a Cauchy sequence, Eq. (\ref{eq:relation_mn_un}) also offers an interesting relation between $\{\Delta^l\}_{l \in \mathbb N}$ and the sequence of interest $\{\mathbf{m}^{l}\}_{l \in \mathbb N}$, which we can use.
Indeed, for any $l_0, m \in \mathbb N$, we can sum Eq. (\ref{eq:relation_mn_un}):
\begin{align}
    \label{eq:telescope_sum}\sum_{l=l_0}^{l_0+m} \Delta^l \geq& \frac{1}{h^2} \sum_{l=l_0}^{l_0+m} u^l \|\mathbf{m}^{l}-\mathbf{m}^{l+1}\| ^2 \\
    \label{eq:take_min_u}\geq& \frac{u^{l_0}}{h^2} \sum_{l=l_0}^{l_0+m} \|\mathbf{m}^{l}-\mathbf{m}^{l+1}\|^2 \\
    \geq& \label{eq:triangle_ineq} \frac{u^{l_0}}{h^2} \|\mathbf{m}^{l_0+m}-\mathbf{m}^{l_0}\|^2
\end{align}
Where Eq. (\ref{eq:take_min_u}) follows because $\{u^l\}_{l \in \mathbb N}$ is strictly increasing, and Eq. (\ref{eq:triangle_ineq}) follows from the triangle inequality. Now, the left-hand side of Eq. (\ref{eq:telescope_sum}) can be reduced to $\sum_{l=l_0}^{l_0+m} \Delta^l= u^{l_0+m+1} - u^{l_0}$. But because we proved in Step 1 that $\{u^l\}_{l \in \mathbb N}$ was a Cauchy sequence, this difference is bounded by a constant. This concludes the proof that $\{\mathbf{m}^{l}\}_{l \in \mathbb N}$ is itself a Cauchy sequence in the Euclidean space.\\

\textbf{Step 3: } We just proved that $\{\mathbf{m}^{l}\}_{l \in \mathbb N}$ was a Cauchy sequence. Therefore $\{\mathbf{m}^{l}\}_{l \in \mathbb N}$ can only converge to a single value $\mathbf{m}^{*}$. We now use the continuity of function $g$ to conclude that $\mathbf{m}^{*}$ has to be a solution of the initial equation~(\ref{eq:condition}):
\begin{align}
    \mathbf{m}^{*} &= \lim_{l \rightarrow \infty} \mathbf{m}^{l+1} = \lim_{l \rightarrow \infty} g(\mathbf{m}^{l}) \\
    &= g(\lim_{l \rightarrow \infty} \mathbf{m}^{l}) = g(\mathbf{m}^{*})
\end{align}

\section{Further details on MTA}
\label{appendix:further_details}
We summarize the traditional mode seeking MeanShift procedure, upon which our approach is based, in Algorithm~\ref{alg:mean_shift}. Moreover, our robust multi-modal MeanShift for test-time augmentation, named MTA, is presented in Algorithm~\ref{alg:robust_mean_shift} in a non-vectorized manner to highlight each operation. The handcrafted prompts~\cite{clip} for ensembling are listed in Table~\ref{tab:handcrafted}. We use $N$=64 augmented views (63 from random cropping (RandomCrop) and the original image) in all our experiments except in Table \ref{tab:augs} to be consistent with DiffTPT which uses 128 augmented views (63 from diffusion, 64 from random cropping and the original image). Table \ref{tab:hp_search} shows the interdependency of $\lambda$ and $\lambda_{\mathbf{y}}$ and the role of the {\em inlierness} scores: as $\lambda_{\mathbf{y}}$ approaches 0, it tends toward a peak selection and trivial solutions; conversely, as $\lambda_{\mathbf{y}}$  grows, it
tends to MeanShift with uniform {\em inlierness} scores. 
\begin{algorithm}
  \caption{Mode seeking MeanShift \cite{comaniciu1999mean}
    \label{alg:mean_shift}}
  \begin{algorithmic}[1]
    \Require{$h>0$ the bandwidth, $K$ a kernel function (e.g., Gaussian kernel), ${\mathbf m}^0$ a first estimate of the mode, a set of data points $({\mathbf f}_p)_{1 \leq p \leq N}$, a threshold value $\epsilon$}
    \State $l$ $\gets$ $0$
      \While{$l = 0$ or $\|{\mathbf m}^l - {\mathbf m}^{l-1}\| \geq \epsilon$}
        \State $\mathbf{m}^{l+1}$ $\gets$ $\frac{\sum_{p = 1}^{N} K(\mathbf{f}_p - \mathbf{m}^l) \mathbf{f}_p}{\sum_{p = 1}^N  K(\mathbf{f}_p - \mathbf{m}^l)}$ \Comment{mode update}
        \State $l$ $\gets$ $l+1$
      \EndWhile
      \State ${\mathbf m}$$\gets$ ${\mathbf m}^{l-1}$
      \State \Return{${\mathbf m}$} 
  \end{algorithmic}
\end{algorithm}
\begin{algorithm*}
\setstretch{1.2}
  \caption{MTA with Gaussian kernel
    \label{alg:robust_mean_shift}}
  \begin{algorithmic}[1]
   \Require{A set of augmented embeddings $({\mathbf f}_p)_{1 \leq p \leq N}$} with ${\mathbf f}_1$ being the original image, a set of class embeddings $({\mathbf t}_k)_{1 \leq k \leq K}$, a threshold value $\epsilon$, $\tau$ the temperature variable of the CLIP model.
    \State $w_{p,q}$ $\gets$ \texttt{Affinity}(${\mathbf f}_p$, ${\mathbf f}_q$, $({\mathbf t}_k)_{1 \leq k \leq K}$, $\tau$) \hspace{0.25cm} $\forall~p,q \in \{1,...,N\}$ \Comment{See Algorithm \ref{alg:affinity}}
    \State $h_p^2$ $\gets$ $\frac{1}{\rho(N-1)} \sum_{q \in I_p} \|f_p-f_q\|^2$  \hspace{0.25cm} $\forall~p \in \{1,...,N\}$ \Comment{$I_p$ the closest neighbors of p, $\rho$ set to $0.3$}

    \State ${\mathbf m}$ $\gets$ ${\mathbf f}_1$  \Comment{mode initialization}
    \State $y_p$ $\gets$ $\frac{1}{N}$ $\forall~p \in \{1,...,N\}$ \Comment{Initial {\em inlierness} scores uniform}
      \While{\textbf{\textit{(1)}} and \textbf{\textit{(2)}} not converged}
      \State $n$ $\gets$ $0$
      \State ${\mathbf y}^0$ $\gets$ ${\mathbf y}$
        \While{$n = 0$ or $\|{\mathbf y}^n - {\mathbf y}^{n-1}\| \geq \epsilon$}
             \State $y_p^{(n+1)}$ $\gets$ $\frac{\exp \left ( ( K({\mathbf f}_p - {\mathbf m}) + \lambda \sum_{q = 1}^N w_{p,q} y_q^{(n)})/ \lambda_{{\mathbf y}}   \right)}{\sum_{j=1}^N \exp \left ( ( K({\mathbf f}_j - {\mathbf m}) + \lambda \sum_{q = 1}^N w_{j,q} y_q^{(n)})/\lambda_{{\mathbf y}} \right)}$   \hspace{0.25cm} $\forall~p \in \{1,...,N\}$ \Comment{\textbf{\textit{(1)}} {\em inlierness} scores update} 
             \State $n$ $\gets$ $n+1$
        \EndWhile
        \State ${\mathbf y}$ $\gets$ ${\mathbf y}^{n-1}$ 
        \State $l$ $\gets$ $0$
        \State ${\mathbf m}^0$ $\gets$ ${\mathbf m}$
        \While{$l = 0$ or $\|{\mathbf m}^l - {\mathbf m}^{l-1}\| \geq \epsilon$}
            \State $\mathbf{m}^{l+1}$ $\gets$ $\frac{\sum_{p = 1}^{N}  y_p K(\mathbf{f}_p - \mathbf{m}^l) \mathbf{f}_p}{\sum_{p = 1}^N y_p  K(\mathbf{f}_p - \mathbf{m}^l)}$ \Comment{\textbf{\textit{(2)}} mode update}
            \State $l$ $\gets$ $l+1$
        \EndWhile
        \State ${\mathbf m}$ $\gets$ ${\mathbf m}^{l-1}$
      \EndWhile
      \State \Return{$\argmax_{k} ~{\mathbf m}^t {\mathbf t}_k$} \Comment{return prediction based on the mode}
  \end{algorithmic}
\end{algorithm*}

\begin{algorithm*}
\setstretch{1.2}
  \caption{Affinity measure based on predictions
    \label{alg:affinity}}
  \begin{algorithmic}[1]
    \Function{Affinity}{${\mathbf f}_p$, ${\mathbf f}_q$, $({\mathbf t}_k)_{1 \leq k \leq K}$, $\tau$}
    \If{$p=q$}
    \State \Return 0
    \EndIf
    \State $l_{p,k}$ $\gets$ $\tau {\mathbf f}_p^t {\mathbf t}_k$ \textbf{;} \hspace{0.1cm} $l_{q,k} $ $\gets$ $ \tau {\mathbf f}_q^t {\mathbf t}_k$ \hspace{0.25cm} $\forall~k \in \{1,...,K\}$ \Comment{similarity with $\texttt{class}_k$}

    \State $s_{p,k} $ $\gets$ $ \frac{\exp l_{p,k}}{\sum_{j=1}^K \exp l_{p,j}}$\textbf{;} \hspace{0.1cm} $s_{q,k} $ $\gets$ $ \frac{\exp l_{q,k}}{\sum_{j=1}^K \exp l_{q,j}}$ \hspace{0.25cm} $\forall~k \in \{1,...,K\}$ \Comment{Softmax operation}
    \State $w_{p,q}$ $\gets$ ${\mathbf s}^t_p {\mathbf s}_q$ 
    \State \Return{$w_{p,q}$}
    \EndFunction
  \end{algorithmic}
\end{algorithm*}

\begin{table}[t]

    \centering
    \caption{Effect of $\lambda$ and $\lambda_{\mathbf{y}}$ on the ImageNet dataset. Reported value is the top-1 accuracy averaged over 3 random seeds.}
    %\vspace{-0.1cm}
    \label{tab:hp_search}
    \resizebox{\linewidth}{!}{
    \begin{tabular}{c|ccc>{\columncolor{LightGray}}cccccccc}
    \toprule
          \diagbox{$\lambda$}{$\lambda_{\mathbf{y}}$}  &  0.01& 0.05 & 0.1 &  0.2 & 0.4 & 0.8 & 1.6 & 3.2 & 10 & 100 & \multicolumn{1}{c}{\makecell{$\rightarrow \infty$ \\ (MeanShift) }}   \\
         \midrule
        0   & 66.7 & 66.7 & 66.8 & 68.3 & 65.8 & 65.3 & 65.6 & 65.9  & 66.0 & 66.1 & 66.1 \\
        0.5   & 66.7 & 66.7 & 66.8 & 68.7 & 67.7 & 66.8  & 66.4 & 66.2 & 66.1&66.1 & - \\  
        1   & 66.7 & 66.7& 66.9 & 68.9 & 68.2 & 67.4 & 66.9 & 66.5 & 66.3&66.1 & - \\
        2   & 66.8 & 66.8 & 67.1 & 69.1 & 68.8 & 68.0 & 67.4 & 67.0 & 66.5&66.1 & - \\
        \rowcolor{LightGray} 4  & 66.6 & 66.5 & 66.9 & \textbf{69.3} & 69.1 & 68.6 &68.0 & 67.5 & 66.8& 66.2 & - \\
        8   & 62.0& 62.5 & 64.2 & 68.7 & 69.3 & 69.0 & 68.5 & 68.1 & 67.2& 66.3& - \\
        16   &57.3& 58.5 & 61.0 & 65.8 & 69.1 & 69.3 & 68.9 & 68.5 & 67.7&66.4 & - \\
    \end{tabular}}
\end{table}

\section{Additional results}
\label{appendix:additional_results}
\paragraph{Zero-shot (Section \ref{sec:zero_shot}).} We report detailed results for Table \ref{tab:zero_shot_imagenet}, Table \ref{tab:zero_shot_datasets} and Table \ref{tab:augs} with average top-1 accuracy and standard deviation in Table \ref{tab:zero_shot_imagenet_details}, Table \ref{tab:zero_shot_datasets_details} and Table \ref{tab:augs_details} respectively.
\begin{table*}[t!]
\caption{Details of Table \ref{tab:zero_shot_imagenet} with averaged top-1 accuracy and standard deviation computed over 3 random seeds.} \label{tab:zero_shot_imagenet_details}
\sisetup{table-format=-1.2}   % 2 decimals, leave 
\centering
\resizebox{0.75\linewidth}{!}{
   \begin{tabular}{llcccccc}
\toprule 
\multicolumn{1}{l}{\bf Method}  
&\multicolumn{1}{c}{}
&\multicolumn{1}{c}{\makecell{\bf ImageNet}}
&\multicolumn{1}{c}{\makecell{\bf -A}} 
&\multicolumn{1}{c}{\makecell{\bf -V2}} 
&\multicolumn{1}{c}{\makecell{\bf -R}}
&\multicolumn{1}{c}{\makecell{\bf -Sketch}}
&\multicolumn{1}{c}{\bf Average} 
\\ \midrule
\multirow{2}{*}{\makecell{TPT}} & \multirow{2}{*}{\makecell{\xmark}}         & 68.94 & 54.63 & 63.41 & 77.04 & 47.97  & 62.40  \\
 &           & $\pm$ .06 & $\pm$ .21 & $\pm$ .12 & $\pm$ .02 & $\pm$  .05 & $\pm$  .03 \\
 \midrule
\multirow{2}{*}{\makecell{MTA}} & \multirow{2}{*}{\makecell{\cmark}}   & 69.29 & 57.41 & 63.61 & 76.92 & 48.58 & 63.16 \\
 &           & $\pm$ .09 & $\pm$ .15 & $\pm$ .07 & $\pm$ .13 & $\pm$  .05 & $\pm$  .07 \\
    \midrule
\multirow{2}{*}{\makecell{MTA + Ensemble}} & \multirow{2}{*}{\makecell{\cmark}}    & 70.08 & 58.06 & 64.24 & 78.33 & 49.61 & 64.06 \\
 &           & $\pm$ .03 & $\pm$ .07 & $\pm$ .09 & $\pm$ .11 & $\pm$  .06 & $\pm$  .06 \\
\midrule 
\midrule 
\multirow{2}{*}{\makecell{TPT + CoOp}} & \multirow{2}{*}{\makecell{\xmark}}         & 73.61 & 57.85 & 66.69 & 77.99 & 49.59     & 65.14 \\
 &           & $\pm$ .17 & $\pm$ .34 & $\pm$ .25 & $\pm$ .69 & $\pm$  .34 & $\pm$  .1 \\
 \midrule
\multirow{2}{*}{\makecell{MTA + CoOp}} & \multirow{2}{*}{\makecell{\cmark}}  & 73.99 & 59.29 & 66.97 & 78.2 & 49.96 & 65.68 \\
  &           & $\pm$ .18 & $\pm$ .12 & $\pm$ .25 & $\pm$ .76 & $\pm$  .46 & $\pm$  .25 \\

      \bottomrule
   \end{tabular}}
\end{table*}

\begin{table*}[t!]
\caption{Details of Table \ref{tab:zero_shot_datasets} with averaged top-1 accuracy and standard deviation computed over 3 random seeds.}
\label{tab:zero_shot_datasets_details}
\centering
\resizebox{\textwidth}{!}{
\begin{tabular}{lccccccccccc}
\toprule
Method & SUN397 & Aircraft & EuroSAT & Cars & Food101 & Pets &  Flower102 & Caltech101 & DTD & UCF101 & Average
\\ \midrule 
\multirow{2}{*}{\makecell{TPT}} & 65.41 & 23.1 & 42.93 & 66.36 & 84.63 & 87.22 & 68.86 & 94.12 & 46.99 & 68.00 & 64.76\\
 & $\pm$ .03 & $\pm$ .39 & $\pm$ .2 & $\pm$ .31 & $\pm$  .03 & $\pm$  .19 & $\pm$ .32 & $\pm$ .21 & $\pm$ .31 & $\pm$ .22 & $\pm$  .05\\
\midrule
\multirow{2}{*}{\makecell{MTA}}  & 64.98 & 25.32 & 38.71 & 68.05 & 84.95 & 88.22 & 68.26 & 94.13 & 45.59 & 68.11 & 64.63 \\
  & $\pm$ 0 & $\pm$ .25 & $\pm$ .22 & $\pm$ .16 & $\pm$  .06 & $\pm$  .07 & $\pm$ .08 & $\pm$ .02 & $\pm$ .18 & $\pm$ .11 & $\pm$  .02\\
\midrule
\multirow{2}{*}{\makecell{MTA + E.}}  & 66.67 & 25.2  & 45.36 & 68.47 & 85.00 & 88.24 & 68.06 & 94.21 & 45.9 & 68.69 & 65.58 \\
 & $\pm$ .05 & $\pm$ .37 & $\pm$ .16 & $\pm$ .08 & $\pm$  .03 & $\pm$  .07 & $\pm$ .2 & $\pm$ .21 & $\pm$ .09 & $\pm$ .15 & $\pm$  .05\\
\bottomrule

\end{tabular}}

\end{table*}

\begin{table*}[t!]
\caption{Details of Table \ref{tab:augs} with averaged top-1 accuracy and standard deviation computed over 3 random seeds.}
\label{tab:augs_details}
\centering
\resizebox{0.7\linewidth}{!}{
\begin{tabular}{llcccccccc}
\toprule
\multicolumn{1}{l}{\bf Augmentation} 
&\multicolumn{1}{l}{\bf Method}  
&\multicolumn{1}{c}{\makecell{\textbf{ImageNet}}}
&\multicolumn{1}{c}{\makecell{\textbf{-A}}} 
&\multicolumn{1}{c}{\makecell{\textbf{-V2}}} 
&\multicolumn{1}{c}{\makecell{\textbf{R}}}
&\multicolumn{1}{c}{\makecell{\textbf{-Sketch}}}
&\multicolumn{1}{c}{\bf Average} 
\\ \midrule

\multirow{4}{*}{\makecell{RandomCrop}} & \multirow{2}{*}{\makecell{TPT}} 
      & 68.15 & 51.23 & 66.17 & 76.88 & 49.31 & 62.35 \\
    &   & $\pm$ .3 & $\pm$ .31 & $\pm$ .2 & $\pm$ .2 & $\pm$  .2 & $\pm$  .05 \\

& \multirow{2}{*}{\makecell{MTA}}   & 69.11 & 55.27 & 65.71 & 77.48 & 50.23 & 63.56 \\
&   & $\pm$ .4 & $\pm$ .15 & $\pm$ .4 & $\pm$ .36 & $\pm$  .4 & $\pm$  .11 \\
\midrule
\multirow{4}{*}{\makecell{Diffusion}} 
& \multirow{2}{*}{\makecell{DiffTPT}}  & 67.83 & 53.43  & 65.18  & 76.85 & 50.2 &    62.7  \\
&   & $\pm$ .23 & $\pm$ .64 & $\pm$ .43 & $\pm$ .11 & $\pm$  .36 & $\pm$  .19 \\
& \multirow{2}{*}{\makecell{MTA}} & 69.18 & 54.5 & 64.81 & 76.82 & 51.09 &63.28     \\
&   & $\pm$ .4 & $\pm$ .31 & $\pm$ .1 & $\pm$ .26 & $\pm$  .4 & $\pm$  .07 \\
\bottomrule

\end{tabular}}
\end{table*}

\begin{figure*}[t]
\centering
\subfloat{\label{sfig:a}\includegraphics[width=.24\textwidth]{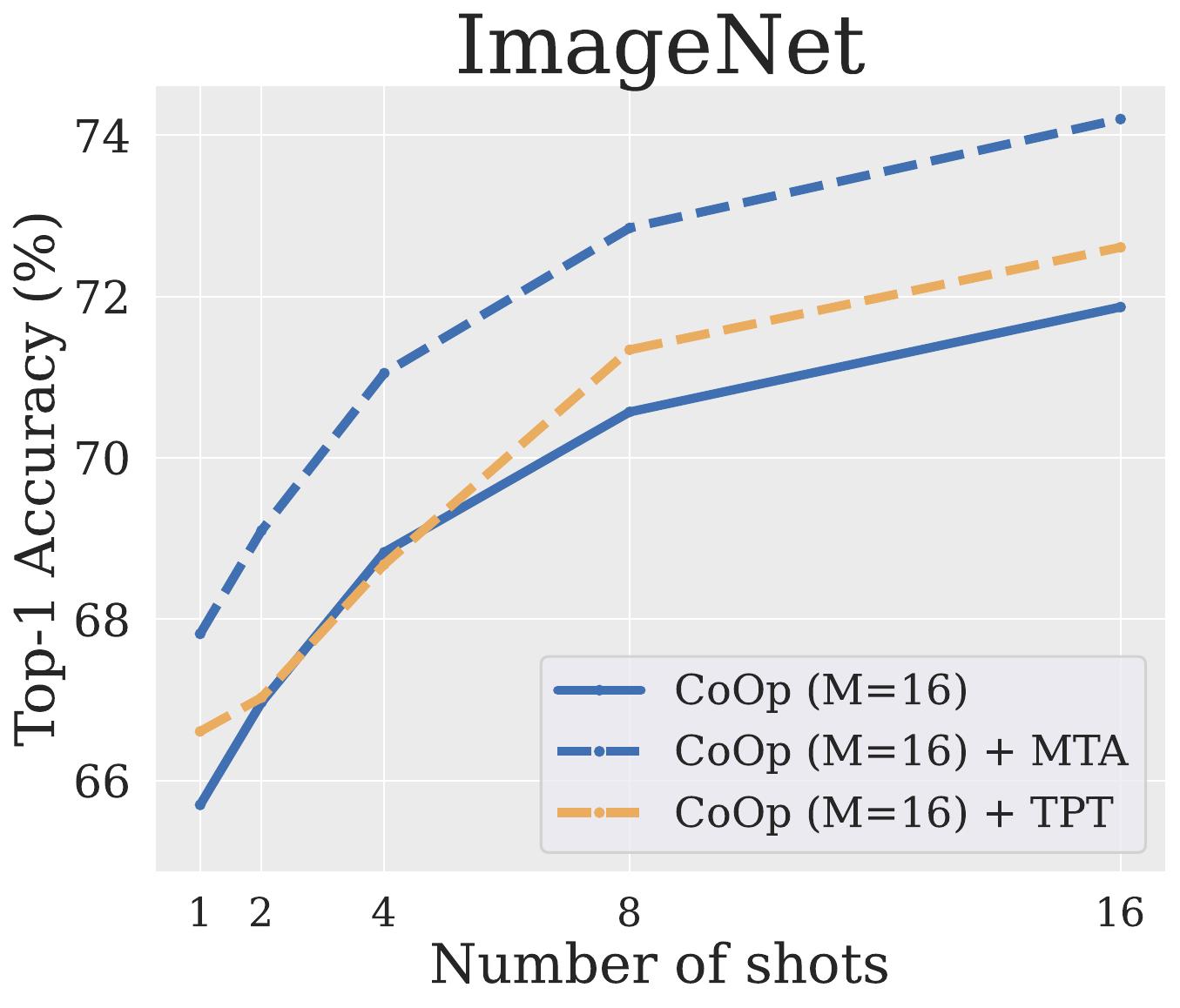}}\hfill
\subfloat{\label{sfig:b}\includegraphics[width=.24\textwidth]{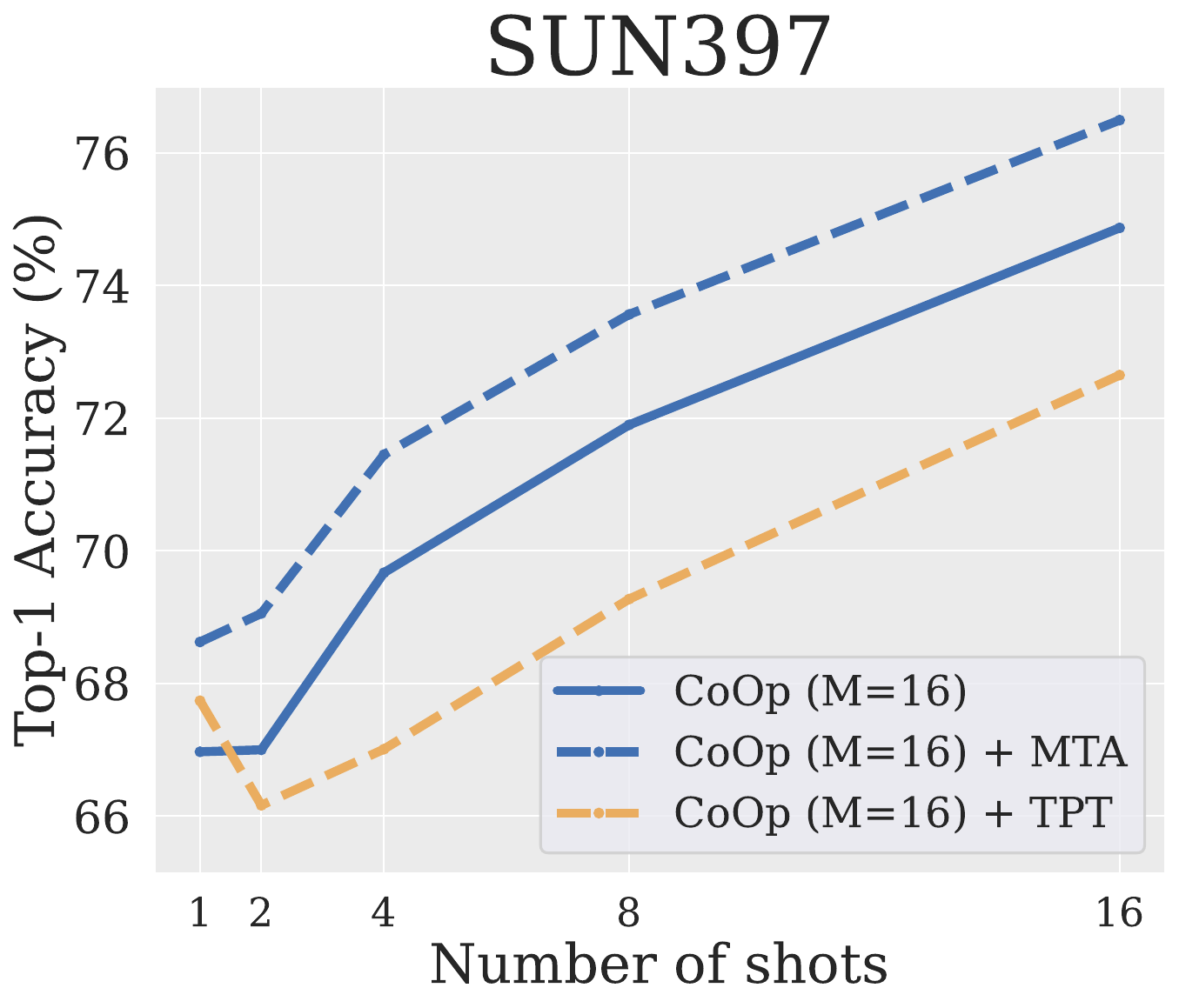}}\hfill
\subfloat{\label{sfig:c}\includegraphics[width=.24\textwidth]{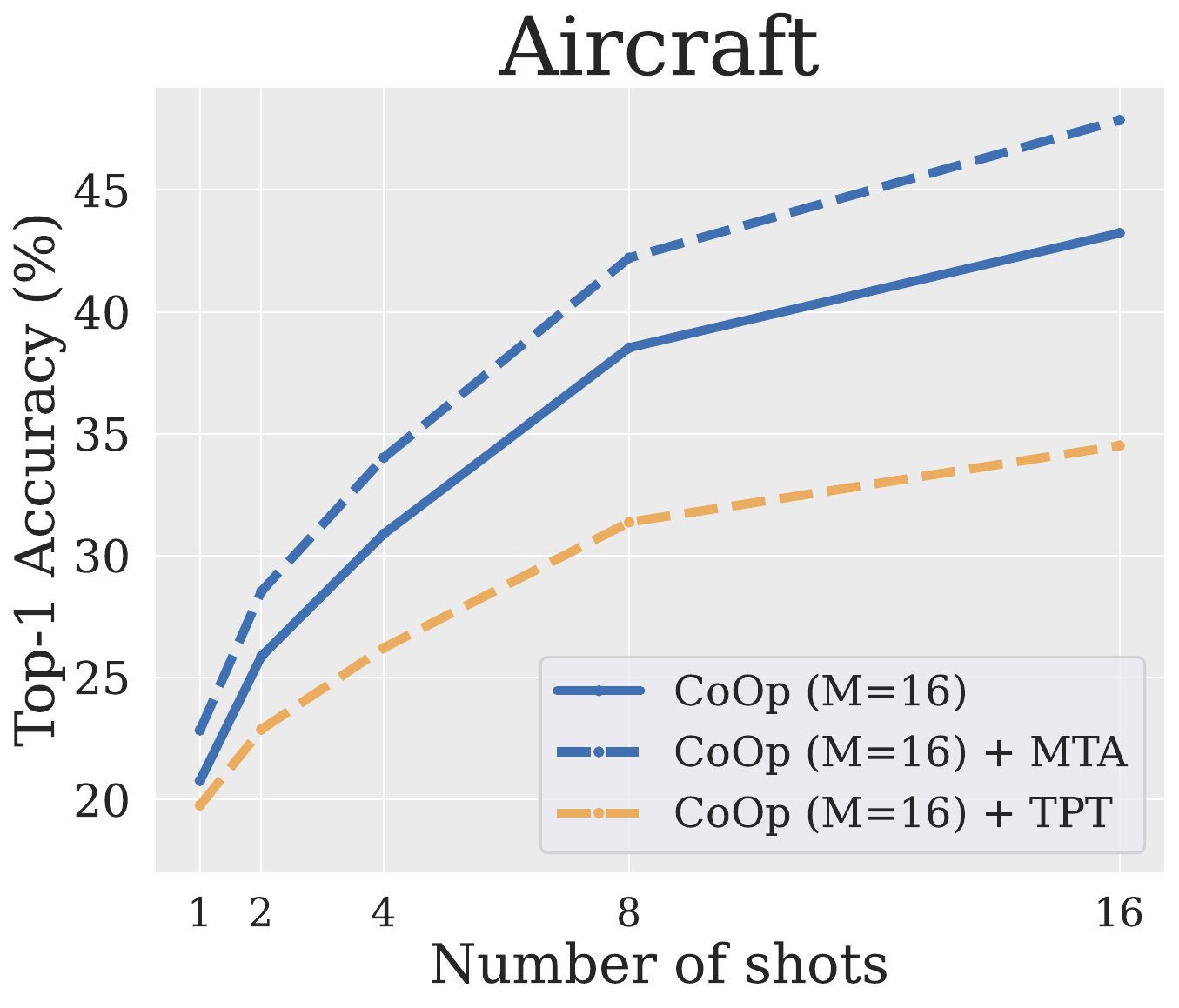}}\hfill
\subfloat{\label{sfig:d}\includegraphics[width=.24\textwidth]{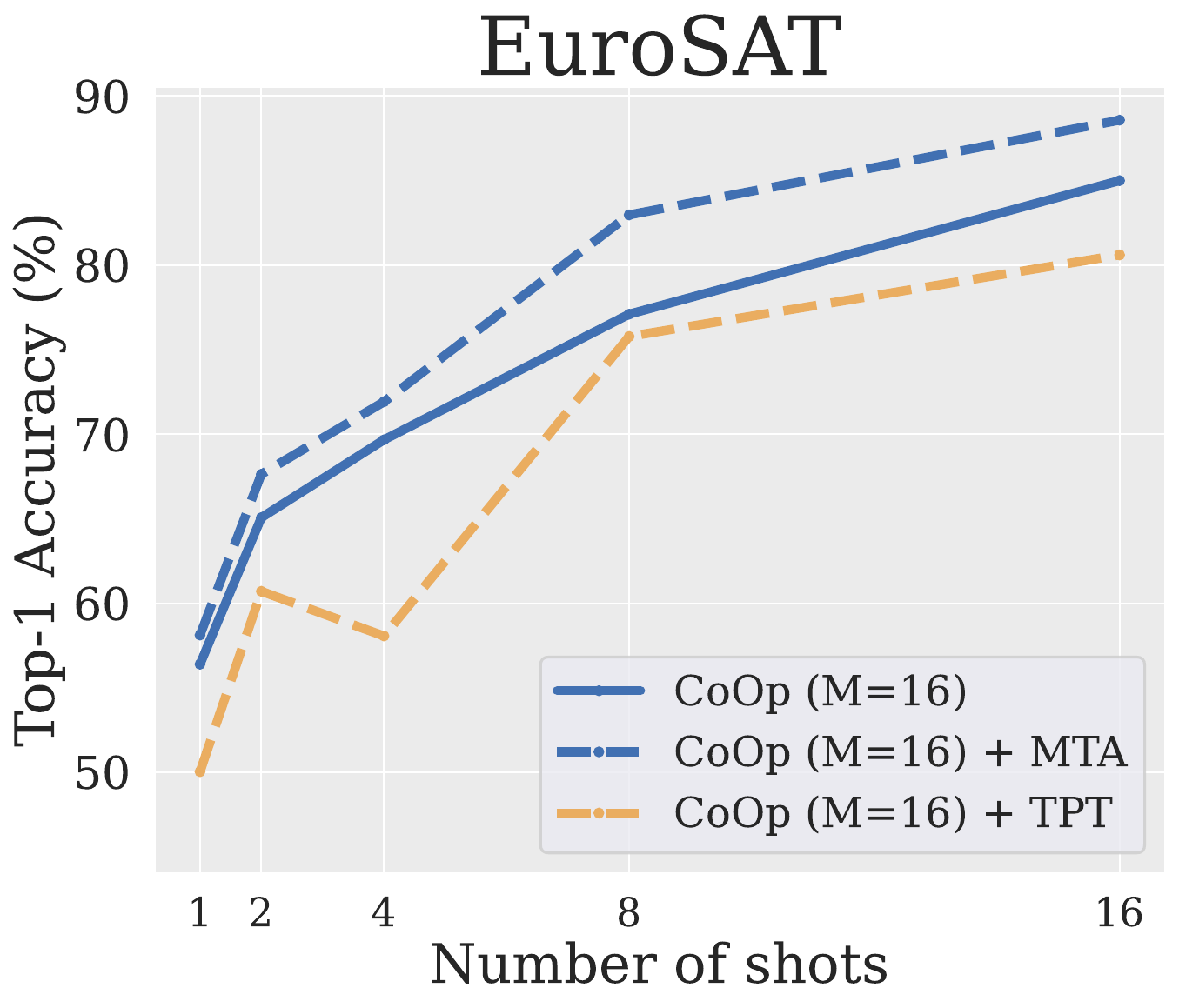}}\hfill
\subfloat{\label{sfig:f}\includegraphics[width=.24\textwidth]{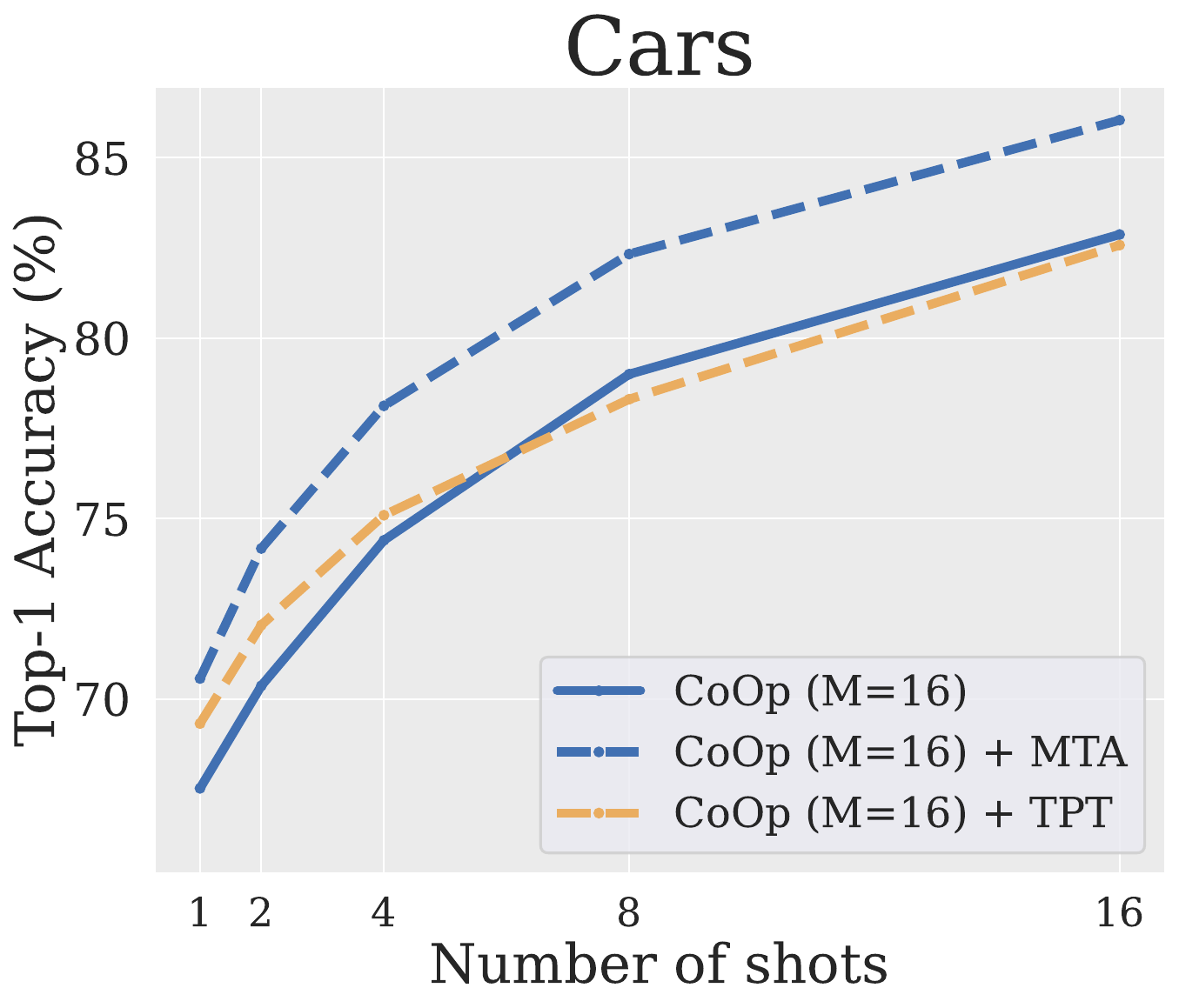}}\hfill
\subfloat{\label{sfig:g}\includegraphics[width=.24\textwidth]{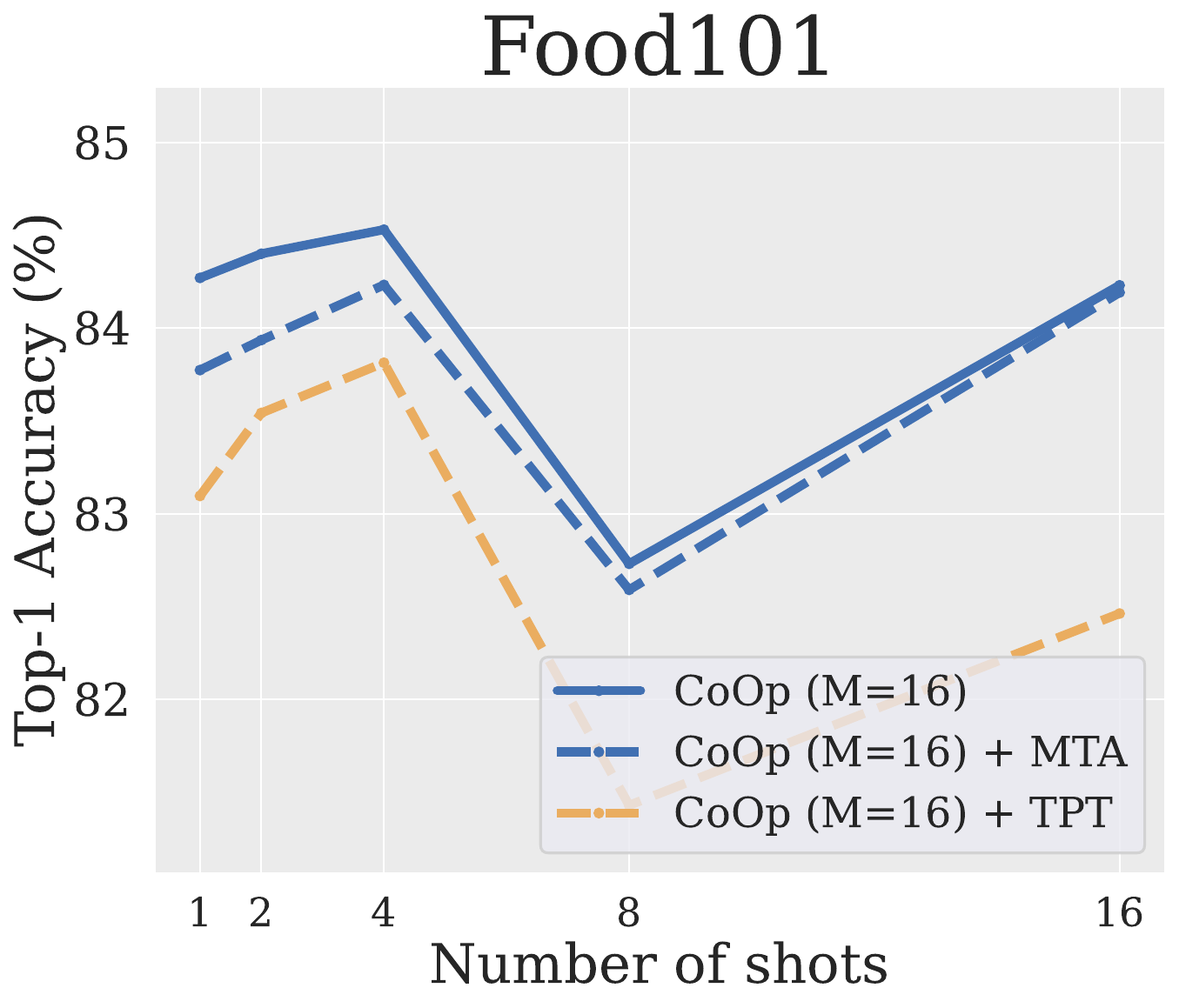}}\hfill
\subfloat{\label{sfig:h}\includegraphics[width=.24\textwidth]{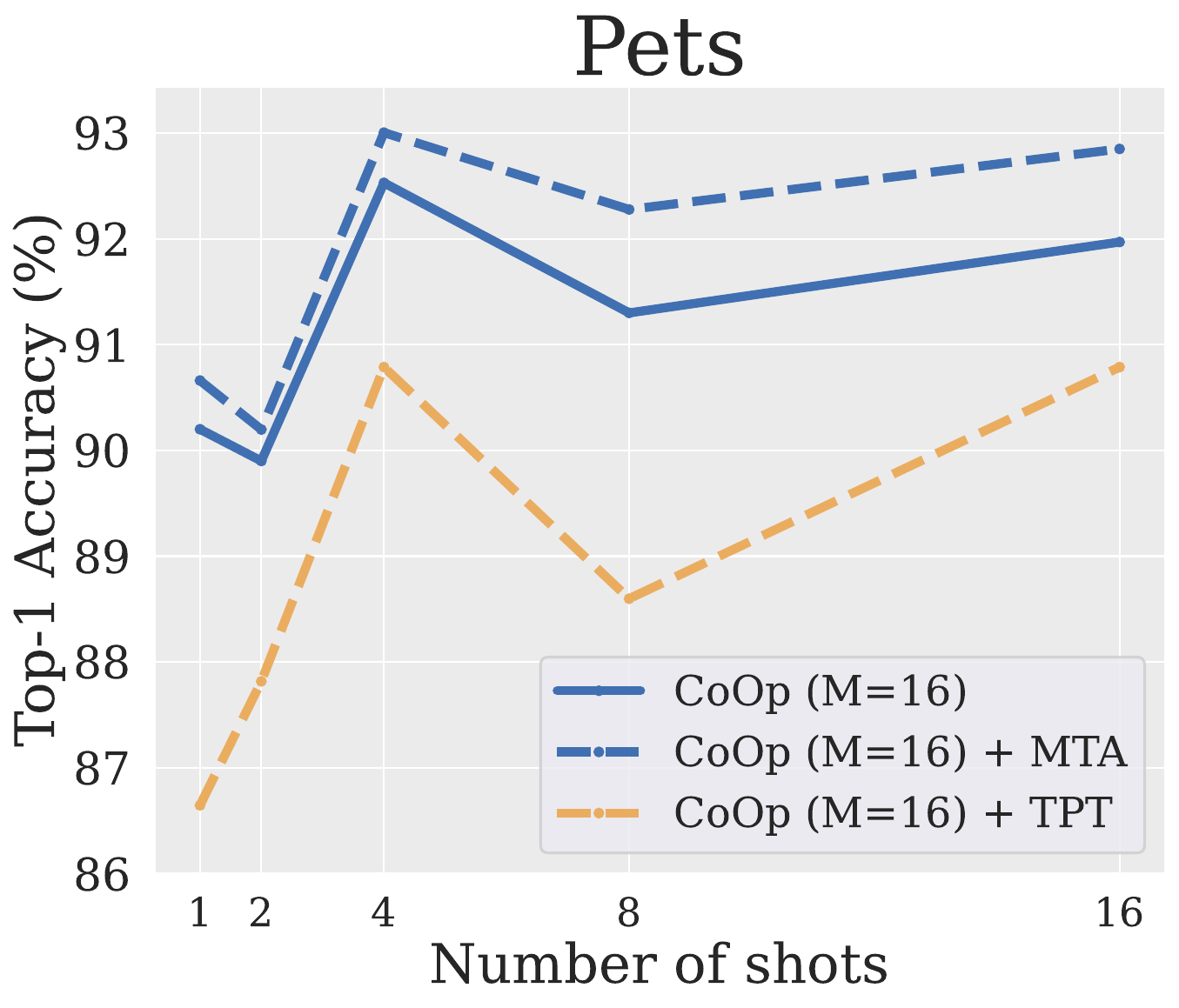}}\hfill
\subfloat{\label{sfig:i}\includegraphics[width=.24\textwidth]{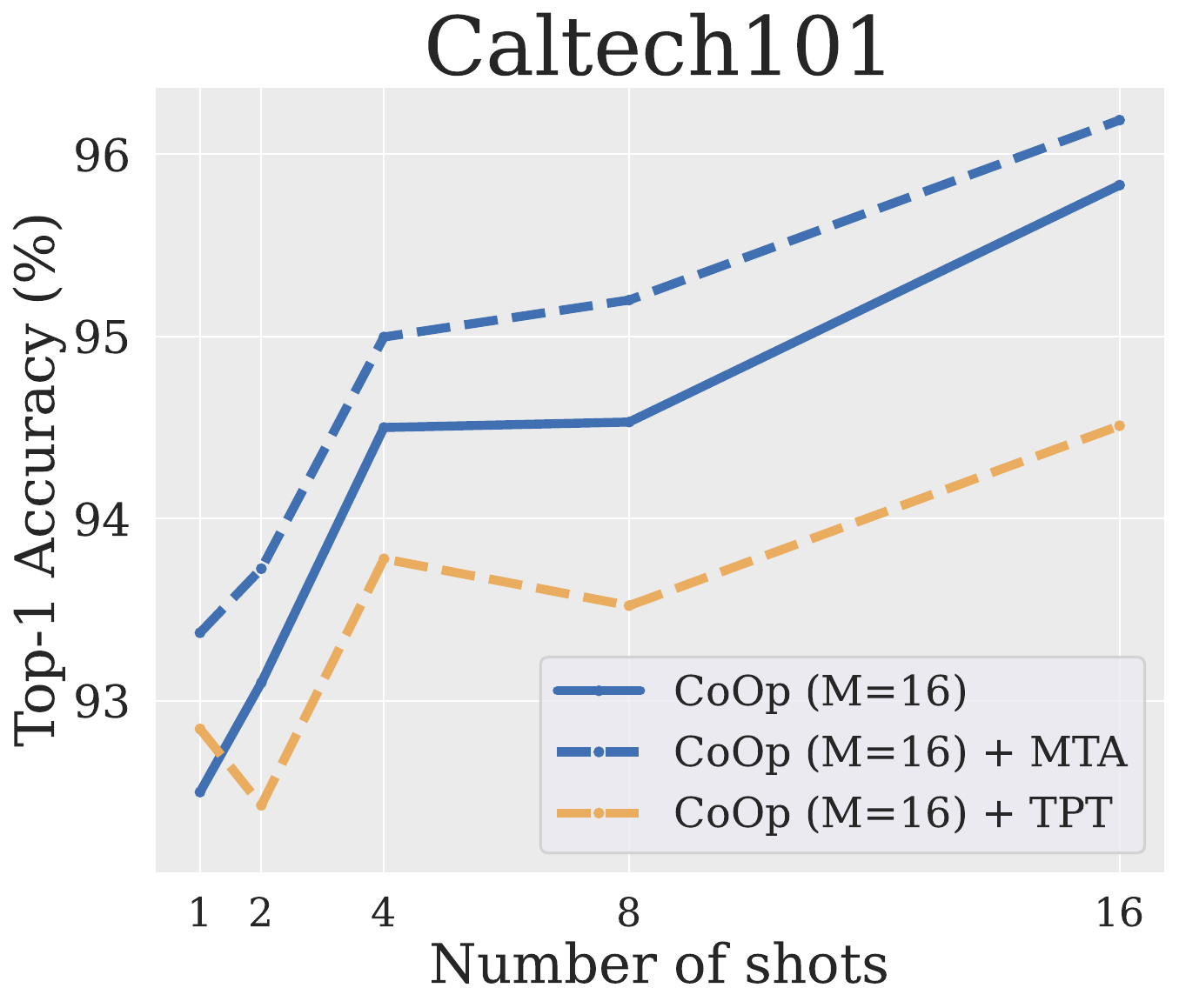}}\hfill
\subfloat{\label{sfig:f}\includegraphics[width=.24\textwidth]{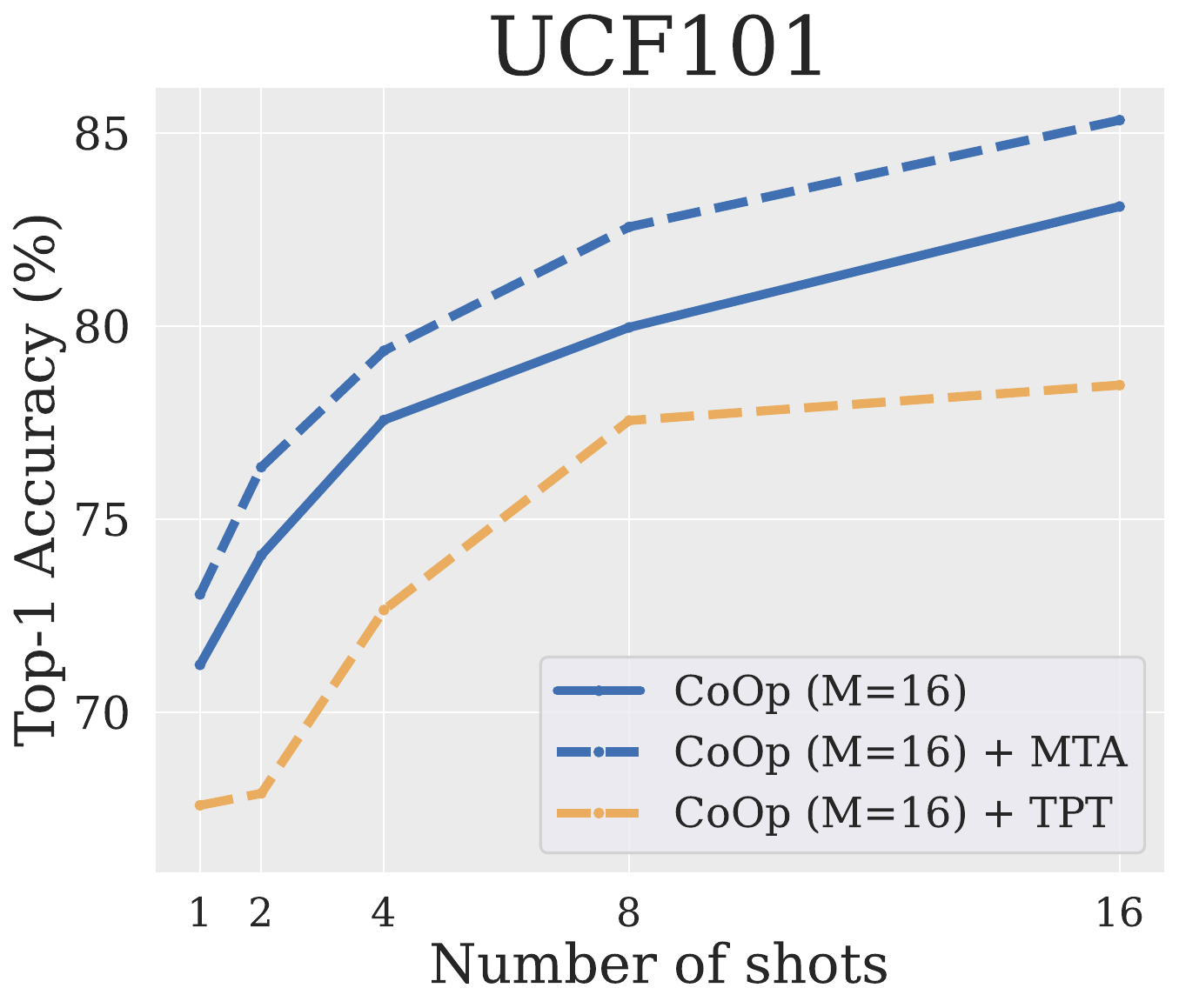}}\hfill
\subfloat{\label{sfig:g}\includegraphics[width=.24\textwidth]{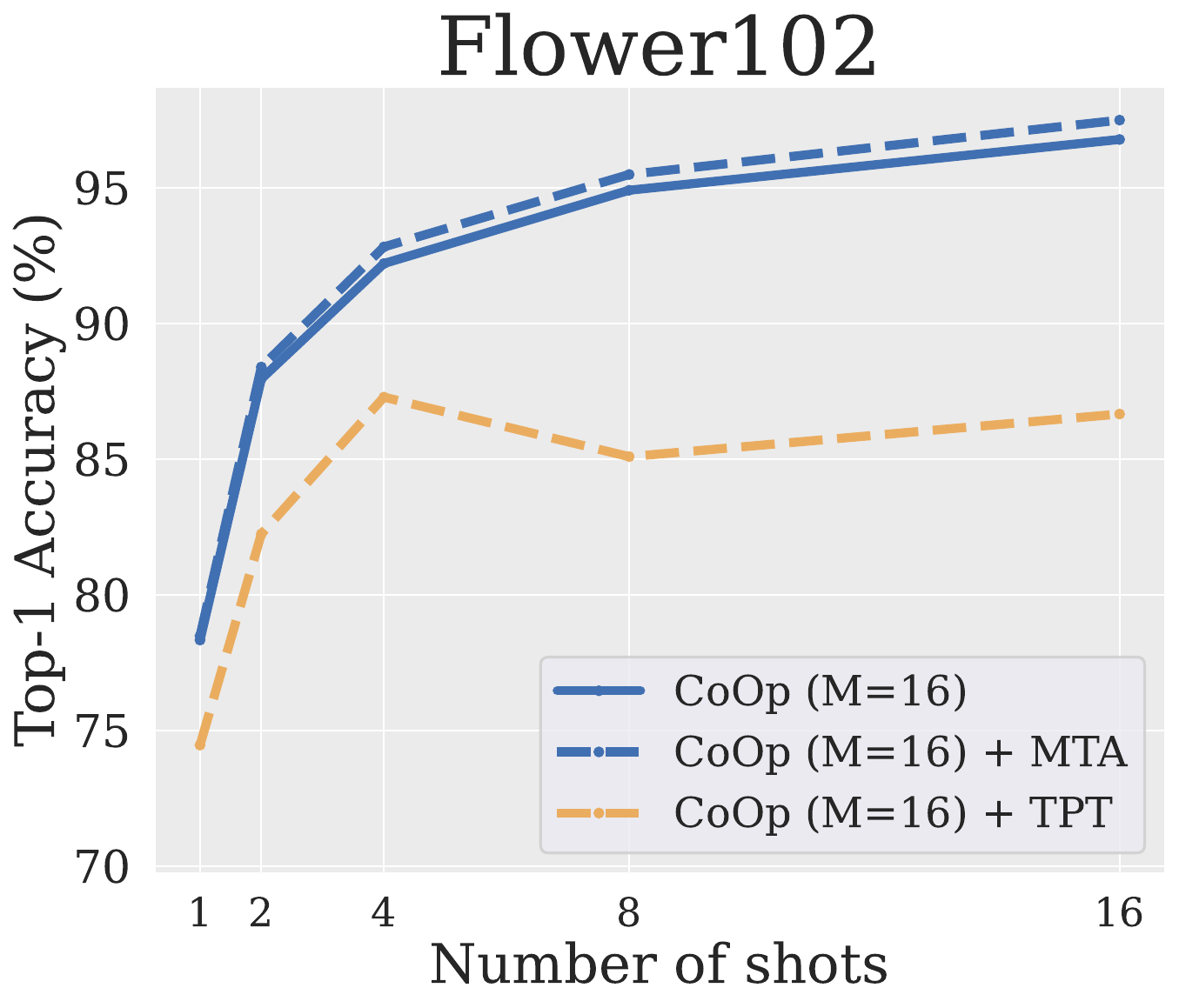}}\hfill
\subfloat{\label{sfig:h}\includegraphics[width=.24\textwidth]{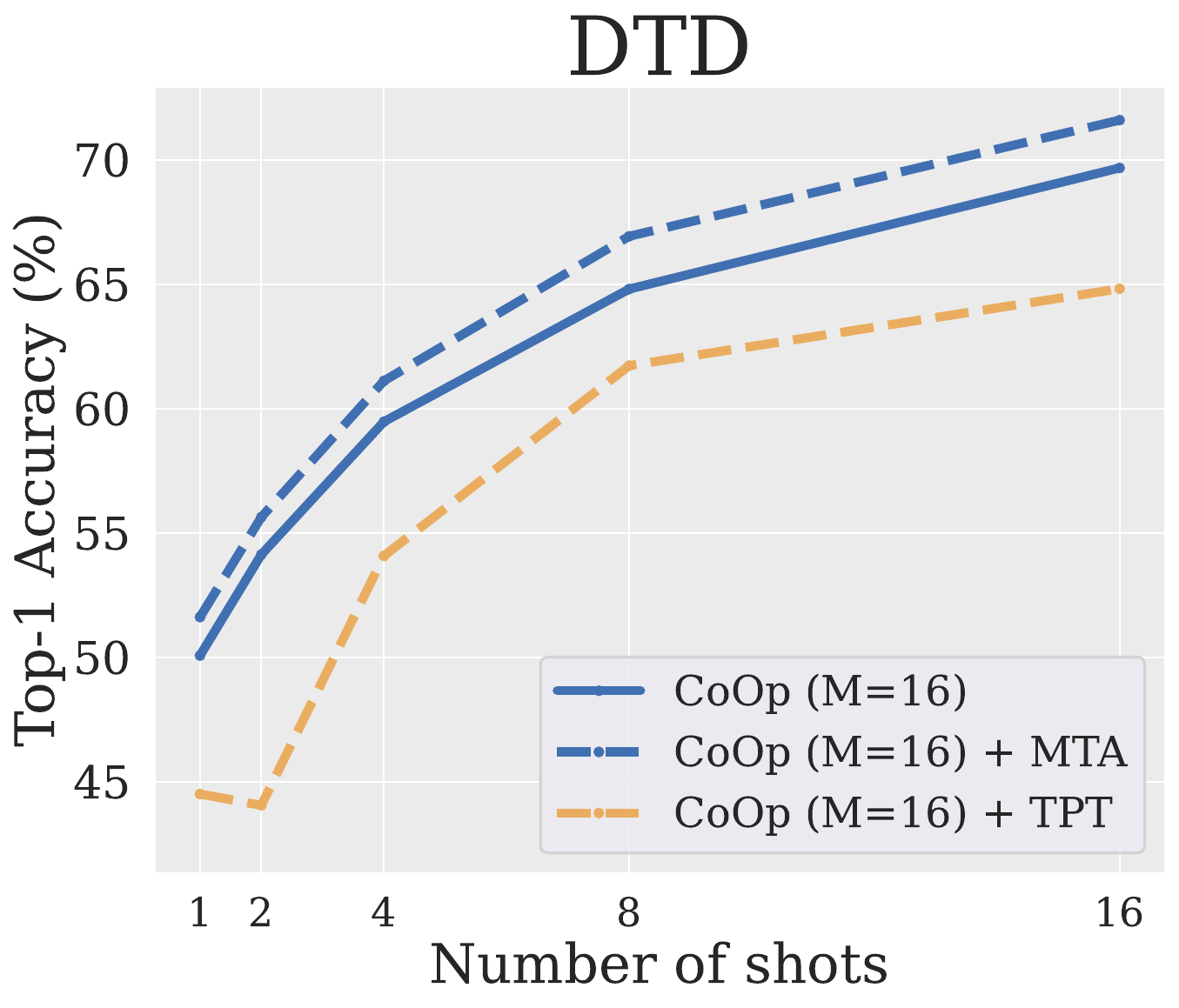}}\hfill
\subfloat{\label{sfig:i}\includegraphics[width=.24\textwidth]{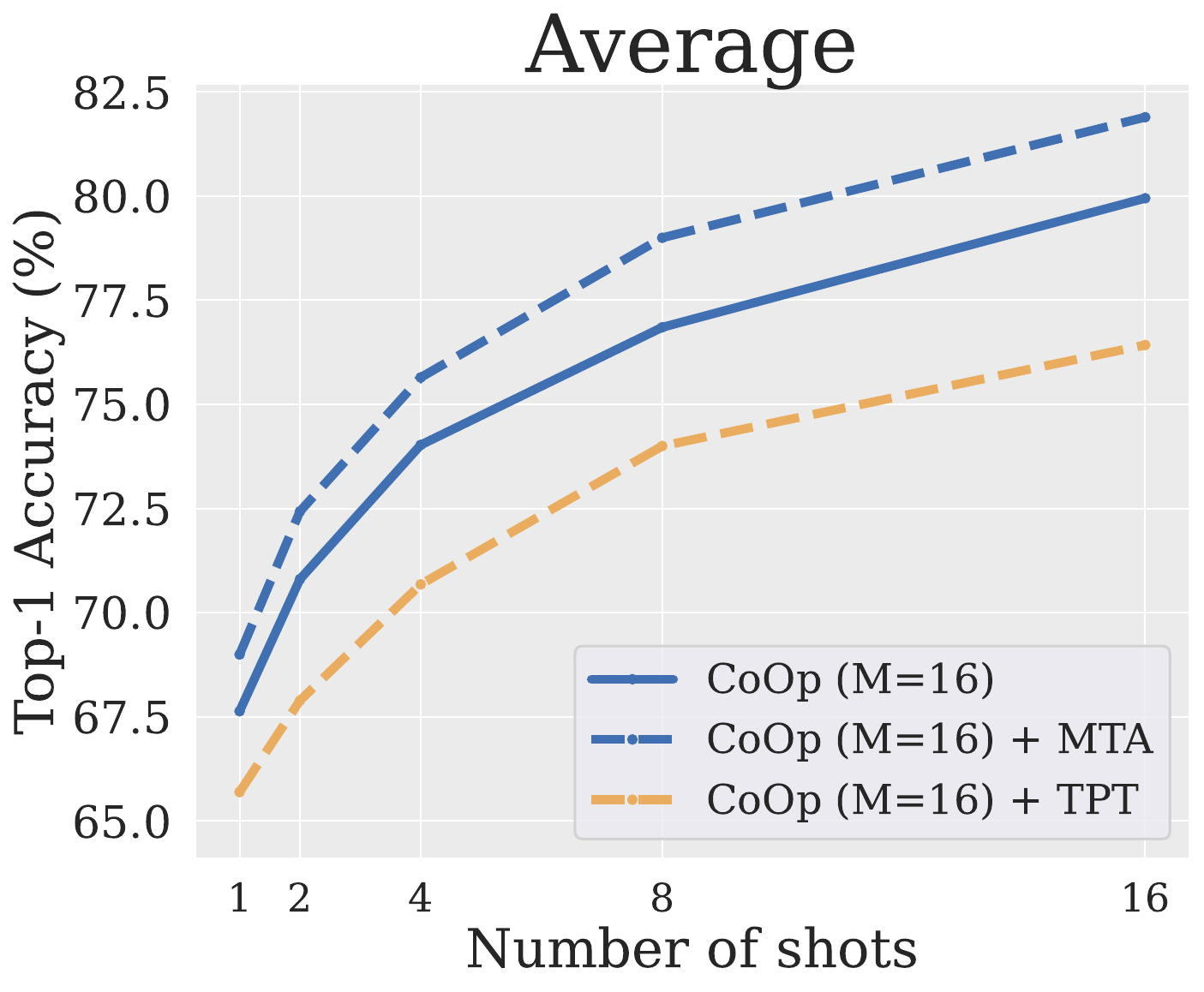}}\hfill
\caption{Additional results for Figure \ref{fig:coop_few_shots} with M=16 tokens for the CoOp pretrained prompts.}
\label{fig:coop_few_shots_16_tokens}
\end{figure*}

\paragraph{Few-shot (Section \ref{sec:few_shot}).} Additional results for CoOp with 16 tokens are depicted in Figure \ref{fig:coop_few_shots_16_tokens}. A similar trend to that shown in Figure \ref{fig:coop_few_shots} is evident, with more pronounced performance degradation observed for TPT. On the contrary, MTA benefits from these more performant prompts.

\paragraph{Ablation study (Section \ref{sec:ablation}).} Details for the 15 datasets for the filtering strategy ablation study of Table \ref{tab:inlier_ablation} are given in Table \ref{tab:inlierness_details_datasets}. With the exception of ImageNet-A, the confidence threshold strategy consistently demonstrates lower performances compared to our {\em inlierness} formulation.

% \begin{table*}[t!]
% \caption{Details of Table \ref{tab:inlier_ablation} for {\em inlierness} score ablation study for the ImageNet and variants.} 
% \label{tab:inlierness_details_imagenets}
% \sisetup{table-format=-1.2}   % 2 decimals, leave 
% \centering
% \resizebox{0.75\linewidth}{!}{
%    \begin{tabular}{lcccccc}
% \multicolumn{1}{l}{\bf Method}  

% &\multicolumn{1}{c}{\makecell{\bf ImageNet}}
% &\multicolumn{1}{c}{\makecell{\bf -A}} 
% &\multicolumn{1}{c}{\makecell{\bf -V2}} 
% &\multicolumn{1}{c}{\makecell{\bf -R}}
% &\multicolumn{1}{c}{\makecell{\bf -Sketch}}
% &\multicolumn{1}{c}{\bf Average} 
% \\ \midrule
% \multirow{2}{*}{\makecell{None (Equal weights)}}      & x & x & x & x & x  & x  \\
%        & $\pm$ x & $\pm$ x & $\pm$ x & $\pm$ x & $\pm$  x & $\pm$  x \\
%  \midrule
% \multirow{2}{*}{\makecell{Confidence thresh. (10\%)}}    & x & x & x & x & x & x \\
%    & $\pm$ x & $\pm$ x & $\pm$ x & $\pm$ x & $\pm$ x & $\pm$  x \\
%     \midrule
% \multirow{2}{*}{\makecell{{\em Inlierness} score}}    & x & x & x & x & x & x \\
%      & $\pm$ x & $\pm$ x & $\pm$ x & $\pm$ x & $\pm$ x & $\pm$  x \\
%       \bottomrule
%    \end{tabular}}
% \end{table*}

\begin{table*}[t!]
\caption{Details of Table \ref{tab:inlier_ablation} for {\em inlierness} scores ablation study. (1) MeanShift (no {\em inlierness} scores) (2) confidence thresh. (10\%) (3) {\em Inlierness} scores. I stands for ImageNet, A for ImageNet-A, V for ImageNet-V2, R for ImageNet-R and K for ImageNet-Sketch. Reported values are averaged top-1 accuracy and standard deviation computed over 3 random seeds.}
\label{tab:inlierness_details_datasets}
\centering
\resizebox{\textwidth}{!}{
\begin{tabular}{lcccccccccccccccc}
\toprule
 & I & A & V & R & K & SUN397 & Aircraft & EuroSAT & Cars & Food101 & Pets &  Flower102 & Caltech101 & DTD & UCF101 & Average
\\ \midrule 
\multirow{2}{*}{\makecell{(1)}} & 66.1 & 48.05 & 60.29 & 67.69 & 40.59 & 63.74 & 25.11 & 24.72 & 66.53 & 83.12 & 85.24 & 66.69 & 91.52 & 44.35 & 65.16 &  59.93 \\
 & $\pm$ .03 & $\pm$ .14 & $\pm$ .23 & $\pm$ .1 & $\pm$  .05 & $\pm$  .09 & $\pm$ .1 & $\pm$ .08 & $\pm$ .2 & $\pm$ .09 & $\pm$  .22 & $\pm$ .25 & $\pm$ .11 & $\pm$ .24 & $\pm$ .05 & $\pm$  .07 \\
\midrule
\multirow{2}{*}{\makecell{(2)}}  & 68.26 & 60.66 & 63.3 & 76.14 & 47.59 & 63.56 & 24.52 & 36.13 & 67.59 & 83.39 & 85.83 & 66.51 & 92.69 & 45.45 & 67.41 & 63.27\\
  & $\pm$ .07 & $\pm$ .19 & $\pm$ .13 & $\pm$ .08 & $\pm$ .05 & $\pm$ .11 & $\pm$ .24 & $\pm$ .04 & $\pm$ .09 & $\pm$ .14 & $\pm$  .32 & $\pm$ .42 & $\pm$ .1 & $\pm$ .1 & $\pm$ .39 & $\pm$  .04 \\
\midrule
\multirow{2}{*}{\makecell{(3)}}  & 69.29 & 57.41 & 63.61 & 76.92 & 48.58  & 64.98 & 25.32 & 38.71 & 68.05 & 84.95 & 88.22 & 68.26 & 94.13 & 45.59 & 68.11 & 64.14\\
 & $\pm$ .09 & $\pm$ .15 & $\pm$ .07 & $\pm$ .13 & $\pm$  .05 & $\pm$ 0 & $\pm$ .25 & $\pm$ .22 & $\pm$ .16 & $\pm$  .06 & $\pm$  .07 & $\pm$ .08 & $\pm$ .02 & $\pm$ .18 & $\pm$ .11 & $\pm$ .01 \\
\bottomrule

\end{tabular}}

\end{table*}

\begin{table*}
\caption{The 80 handcrafted prompts used for majority vote.}
\label{tab:handcrafted}
        "\texttt{a photo of a} [].", 
    "\texttt{a bad photo of a} [].",
    "\texttt{a photo of many} [].",
    "\texttt{a sculpture of a} [].",\\
    "\texttt{a photo of the hard to see} [].",
    "\texttt{a low resolution photo of the} [].",
    "\texttt{a rendering of a} [].",
    "\texttt{graffiti of a} [].",
    "\texttt{a bad photo of the} [].",
    "\texttt{a cropped photo of the} [].",,
    "\texttt{a tattoo of a} [].",
    "\texttt{the embroidered} [].",
    "\texttt{a photo of a hard to see} [].",
    "\texttt{a bright photo of a} [].",\\
    "\texttt{a photo of a clean} [].",
    "\texttt{a photo of a dirty} [].",
    "\texttt{a dark photo of the} [].",\\
    "\texttt{a drawing of a} [].",
    "\texttt{a photo of my} [].",
    "\texttt{the plastic} [].",
    "\texttt{a photo of the cool} [].",\\
    "\texttt{a close-up photo of a} [].",
    "\texttt{a black and white photo of the} [].",
    "\texttt{a painting of the} [].",\\
    "\texttt{a painting of a} [].",
    "\texttt{a pixelated photo of the} [].",
    "\texttt{a sculpture of the} [].",\\
    "\texttt{a bright photo of the} [].",
    "\texttt{a cropped photo of a} [].",
    "\texttt{a plastic} [].",\\
    "\texttt{a photo of the dirty} [].",
    "\texttt{a jpeg corrupted photo of a} [].",
    "\texttt{a blurry photo of the} [].",\\
    "\texttt{a photo of the} [].",
    "\texttt{a good photo of the} [].",
    "\texttt{a rendering of the} [].",\\
    "\texttt{a} [] \texttt{in a video game}.",
    "\texttt{a photo of one} [].",
    "\texttt{a doodle of a} [].", \\
    "\texttt{a close-up photo of the} [].",
    "\texttt{the origami} [].",
    "\texttt{the} [] \texttt{in a video game}.",\\
    "\texttt{a sketch of a} [].",
    "\texttt{a doodle of the} [].",
    "\texttt{a origami} [].",
    "\texttt{a low resolution photo of a} [].",
    "\texttt{the toy} [].",
    "\texttt{a rendition of the} [].",
    "\texttt{a photo of the clean} [].",
    "\texttt{a photo of a large} [].",\\
    "\texttt{a rendition of a} [].",
    "\texttt{a photo of a nice} [].",
    "\texttt{a photo of a weird} [].",\\
    "\texttt{a blurry photo of a} [].",
    "\texttt{a cartoon} [].",
    "\texttt{art of a} [].",
    "\texttt{a sketch of the} [].",\\
    "\texttt{a embroidered} [].",
    "\texttt{a pixelated photo of a} [].",
    "\texttt{itap of the} [].",\\
    "\texttt{a jpeg corrupted photo of the} [].",
    "\texttt{a good photo of a} [].",
    "\texttt{a plushie} [].",\\
    "\texttt{a photo of the nice} [].",
    "\texttt{a photo of the small} [].",
    "\texttt{a photo of the weird} [].",\\
    "\texttt{the cartoon} [].",
    "\texttt{art of the} [].",
    "\texttt{a drawing of the} [].",
    "\texttt{a photo of the large} [].",\\
    "\texttt{a black and white photo of a} [].",
    "\texttt{the plushie} [].",
    "\texttt{a dark photo of a} [].",
    "\texttt{itap of a} [].",
    "\texttt{graffiti of the} [].",
    "\texttt{a toy} [].",
    "\texttt{itap of my} [].",
    "\texttt{a photo of a cool} [].",\\
    "\texttt{a photo of a small} [].",
    "\texttt{a tattoo of the} []."

\end{table*}

\end{document}